\documentclass{article}

\PassOptionsToPackage{numbers, compress}{natbib}
\usepackage{microtype}
\usepackage{graphicx}
\usepackage{subcaption}
\usepackage{booktabs} 

\usepackage{xspace}
\newcommand{\seer}{SeerAttention\xspace}
\newcommand{\seerR}{SeerAttention-R\xspace}
\newcommand{\gate}{AttnGate\xspace}

\newcommand{\Figref}[1]{Figure~{\ref{#1}}}
\newcommand{\tabref}[1]{Table~{\ref{#1}}}

\usepackage{multirow}
\usepackage{graphicx}
\usepackage{subcaption}
\usepackage{xcolor}
\usepackage{siunitx}

\usepackage{comment}

\usepackage{xspace}
\usepackage{multirow}
\usepackage{graphicx}
\usepackage{subcaption}
\usepackage{xcolor}
\usepackage{amsmath}
\usepackage{url}
\usepackage[linesnumbered,ruled,vlined]{algorithm2e}
\usepackage{algorithmic}
\SetKwInOut{Input}{Input}
\SetKwInOut{Output}{Output}

\usepackage[preprint]{neurips_2025}

\usepackage[utf8]{inputenc} 
\usepackage[T1]{fontenc}    
\usepackage{hyperref}       
\usepackage{url}            
\usepackage{booktabs}       
\usepackage{amsfonts}       
\usepackage{nicefrac}       
\usepackage{microtype}      
\usepackage{xcolor}         
\usepackage{wrapfig}

\title{SeerAttention-R: Sparse Attention Adaptation for Long Reasoning}

\author{
Yizhao Gao\thanks{~Equal contribution. $\diamond$ Corresponding author.}$~~^{12}$~~~~~
Shuming Guo\footnotemark[1]$~~^{13}$~~~~
Shijie Cao$^{1}$$^{\diamond}$ ~~~~~
Yuqing Xia$^{1}$~~~~~
Yu Cheng$^{14}$~~~~~
\\
~\bf
Lei Wang$^{14}$~~~~~
Lingxiao Ma$^{1}$~~~~~
Yutao Sun$^{15}$~~~~~
Tianzhu Ye$^{15}$~~~~~
Li Dong$^{1}$~~~~~ \\
~\bf 
Hayden Kwok-Hay So$^{2}$~~~~~
Yu Hua$^{3}$~~~~~
Ting Cao$^{1}$~~~~~
Fan Yang$^{1}$~~~~~
Mao Yang$^{1}$~~~~~
\\\\
~$^1$ Microsoft Research 
~$^2$ The University of Hong Kong \\
~$^3$ Huazhong University of Science and Technology \\
~$^4$ Peking University
~$^5$ Tsinghua University \\
\\
}

\begin{document}

\maketitle

\begin{abstract}

We introduce \seerR, a sparse attention framework specifically tailored for the long decoding of reasoning models.
Extended from \seer,
\seerR retains the design of learning attention sparsity through a self-distilled gating mechanism, while removing query pooling to accommodate auto-regressive decoding. 
With a \textbf{lightweight} plug-in gating, \seerR is \textbf{flexible} and can be easily integrated into existing pretrained model without modifying the original parameters. 
We demonstrate that \seerR, trained on just 0.4B tokens, maintains near-lossless reasoning accuracy with 4K token budget in AIME benchmark under large sparse attention block sizes (64/128).
Using TileLang, we develop a highly optimized sparse decoding kernel that achieves near-theoretical speedups of up to 9x over FlashAttention-3 on H100 GPU at 90\% sparsity.
Code is available at: \url{https://github.com/microsoft/SeerAttention}.

\end{abstract}

\section{Introduction}

Recent reasoning-focused models such as OpenAI o1~\cite{o1}, DeepSeek-R1~\cite{r1}, and Qwen3~\cite{yang2025qwen3} demonstrate that models' capabilities improve significantly through test-time scaling.
By generating longer sequences during inference, these models are able to think and reason more effectively before producing an answer.
Empirically, longer generations correlate with stronger reasoning performance.
For instance, Qwen3-14B~\cite{yang2025qwen3} outperforms DeepSeek-R1-Distill-Qwen-14B~\cite{r1} while producing longer responses on average.
Similarly, harder benchmarks such as AIME24~\cite{aime} require more tokens per generation than easier ones like MATH-500~\cite{math500}.

However, deeper reasoning introduces increasing efficiency challenges.
Due to the auto-regressive nature of decoding, later tokens must attend to a longer context, increasing compute and memory demands for the KV cache.
As a result, the per-token generation cost grows linearly, while the overall generation cost increases quadratically.

Sparse attention offers a promising approach to addressing the long-sequence efficiency challenges.
While it has been studied in general language modeling, its application to reasoning models, which require prolonged decoding, remains underexplored.
Our experiment using oracle sparsity (Section~\ref{subsec:oracle}) shows that attention in reasoning models is also inherently sparse, activating only a subset of important tokens is sufficient to maintain the model's reasoning capability. The key challenge lies in effectively identifying and leveraging this intrinsic sparsity.

In this work, we extend SeerAttention~\cite{seerattn_v1} to SeerAttention-R, a sparse attention framework aimed to improve the long decoding efficiency of reasoning models. 
SeerAttention was originally designed to improve prefill efficiency by selectively activating important attention blocks through a lightweight, self-distilled attention gating mechanism at post-training time. 
SeerAttention-R retains the core design of self-ditilled attention sparsity and introduces modifications to support efficient decoding.
Specifically, it removes sequence-level pooling of query to accommodate auto-regressive decoding and adopts a shared sparsity design aligned with Grouped Query Attention (GQA) to enhance hardware efficiency. 
\seerR can be integrated into any standard transformer-based pretrained model by adding the learnable gate to the attention layer, without fine-tuning original model parameters.

We apply \seerR to multiple reasoning-focused open-source models, including Qwen3-4B, 8B, 14B~\cite{yang2025qwen3} and DeepSeek-R1-Distill-Qwen-14B~\cite{r1}, and evaluate them on several reasoning benchmarks: AIME24, AIME25~\cite{aime}, MATH-500~\cite{math500}, and GPQA-Diamond~\cite{gpqa}.
Since \seerR only requires training the gating module, the distillation is lightweight with just 0.4B tokens from OpenR1-MATH-220K~\cite{openr1} being sufficient.
Across all models and tasks, \seerR consistently outperforms the Quest~\cite{quest} baseline and maintains near-lossless accuracy under a 4k token budget. 
Notably, the accuracy gap further diminishes as model size increases. 
More importantly, this learnable approach enables more \textbf{coarse-grained sparse attention} (e.g., a block size of 64 or 128), which further reduces the overhead from sparse attention scheme and improve hardware efficiency.

We implement the block sparse flash decoding kernel using both TileLang~\cite{tilelang} and Triton~\cite{triton}, and benchmark it on an H100 GPU with FlashAttention-3 (FA3)~\cite{flash3} as the baseline.
Across a range of combination of sequence lengths, batch sizes, and sparsity levels, our TileLang-based kernel consistently outperforms both Triton and FA3.
The gains are especially pronounced at large sequence lengths and batch sizes.
For example, at batch size 16 and sequence length $\geq$ 32k, our TileLang kernel achieves near-theoretical speedups of up to $8.6\times$ at 90\% sparsity over the FA3 baseline, and delivers a $1.7\times$ speedup compared to the Triton counterpart.

\section{SeerAttention-R}

\subsection{A Recap of \seer}
\begin{figure}[h]
    \centering
    \includegraphics[width=1\linewidth]{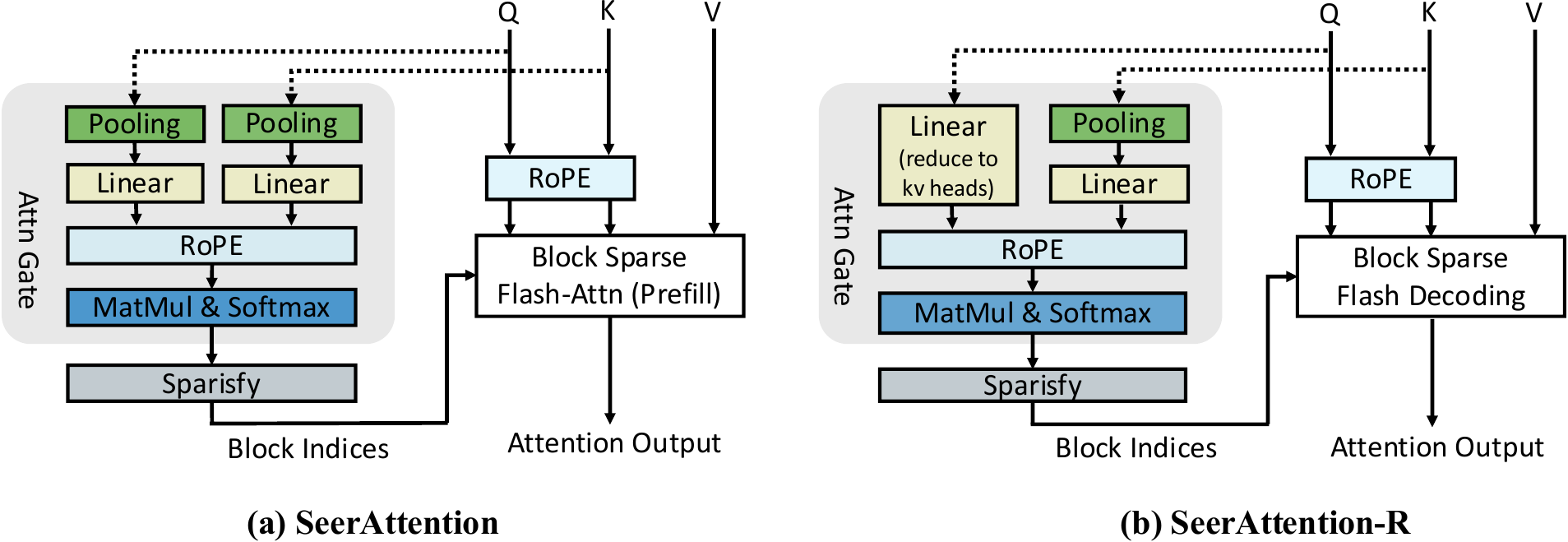}
    \caption{SeerAttention (Sparse Prefill) and SeerAttention-R (Sparse Decode). In \seerR, no sequence dimension compression/pooling operation is applied in Query (Q). Given that modern architectures predominantly use GQA, a linear layer projects the Q from its original number of heads down to the number of KV heads, enabling shared sparsity selection in a GQA group.}
    \label{fig:arch}
\end{figure}

\seer~\cite{seerattn_v1} introduces self-distilled \textit{Attention Gate} (\gate) that dynamically activates sparse blocks in attention computation for efficient long-context \textbf{prefilling}. \Figref{fig:arch}a shows the \gate architecture of \seer, where $\mathbf{Q}$, $\mathbf{K}$ tensors are both compressed (pooled) in the sequence dimension per block number of tokens. The compressed $\mathbf{Q}$, $\mathbf{K}$ tensors are then passed through two newly added linear layers, which serve as learnable parameters in the \gate. With the following positional embedding, matrix-multiplication and softmax operation similar to standard attention, the \gate then generates the 2D block-level attention score estimation. Based on the output, we can selectively activate blocks with higher scores while skipping the rest. 

In the distillation process, the \gate are trained to mimic the 2D block sparse distribution using the ground truth generated by the original pretrained model. This self-distillation training is efficient as the original model weights are frozen. In this way, it brings accurate sparse attention to pretrained full-attention models without costly fine-tuning or pre-training. 
Powered by customized block-sparse flash attention kernels, \seer achieves supreme accuracy-efficiency tradeoff in downstream long-context benchmarks.

\subsection{\seerR: AttnGate for Sparse Decoding}

This work introduces \textbf{\seerR}, an extension of \seer tailored for the long-decoding phase of reasoning models. The foremost difference of \gate design in \seerR is that it does not apply compression/pooling in the sequence dimension of $\mathbf{Q}$ to accommodate the token-by-token auto-regressive decoding process (shown in \Figref{fig:arch}b). 

\begin{subequations}
\footnotesize
\label{eq:gate}
\begin{align}
\mathbf{Q_{gate}} 
  &= \mathrm{RoPE}\Bigl(
       \mathbf{W_{gate}^q} \ \operatorname{reshape}(\mathbf{Q_{nope}}, [..., g\cdot d])
     \Bigr), \\[1ex]
\mathbf{K_{gate}} 
  &= \mathrm{RoPE}
  \Bigl(
       \mathbf{W_{gate}^k} \ \operatorname{concat}[\operatorname{P_{max}}(\mathbf{K_{nope}}),\operatorname{P_{min}}(\mathbf{K_{nope}}),\operatorname{P_{avg}}(\mathbf{K_{nope}})]
     \Bigr), \\[1ex]
\mathbf{S}
  &= \operatorname{softmax}(
       \mathbf{Q_{gate}}\,\mathbf{K_{gate}}^{\!\top}/\sqrt{d_{gate}}
     ).
\end{align}
\end{subequations}
where, $\text{P}_\text{max}$, $\text{P}_\text{min}$, and $\text{P}_\text{avg}$ stand for Max, Min and Average Pooling in sequence dimension, and $g$ is the group size of GQA setting. $d$ and $d_{gate}$ are the hidden dimension of the original model and \gate for each head, respectively. $\mathbf{S}$ is the output score of each block from \gate. The detailed design are discussed as follows.  

\paragraph{Aggregation of Query Heads for Shared Sparsity in GQA}

Group Query Attention (GQA)~\cite{gqa} is widely used in LLMs to reduce KV cache size. In GQA, the query heads are organized into groups, and each group shares a key-value head. Recent sparse attention works SAAP~\cite{shared_sparse_gqa} and NSA~\cite{nsa} show that using identical attention sparsity choices for all queries in a group can improve the efficiency while achieving similar or better performance. In \seerR, we follow this practice and use an linear layer in the $\mathbf{Q}$ branch of \gate to reduce each subgroup of queries to one single head. For example, with 32 query heads and 8 key-value heads (group size $g=4$), there will be 8 sets of linear weights in shape [$d_{gate}$,  $4 \times d$] applying on each group of queries heads, resulting only 8 heads of $\mathbf{Q_{gate}}$. Since we keep the number of heads untouched in $\mathbf{K}$ branch of \gate, the final output of \gate will be key-value heads, achieving a shared decision of sparsity in a group. 

\paragraph{Pooling-based Compression of K}
We follow the practice of \seer that uses pooling operations to compress the sequence dimension of $\mathbf{K}$. The kernel and stride size of pooling are both equal to block size, which can also be understood as non-overlapping chunk-level pooling. To mitigate the potential information loss associated with pooling operations, we employ a composition of Max, Min, and Average pooling operations. The outputs from these pooling operations are concatenated prior to being fed into the subsequent linear layer, similar to \seer. The intuition behind this approach is that Max and Min Pooling can effectively capture outlier values, while Average Pooling helps to keep the overall distribution intact.

\paragraph{Positional Embedding in \gate}
In line with \seer, the decode \gate utilizes the pre-rope $\mathbf{Q}$ and $\mathbf{K}$ 
 tensors as inputs and reapplies RoPE~\cite{rope} within \gate. Given that the 
 branch is compressed along the sequence dimension, the position index is assigned to the initial token of each block.
 In our experiment, we found that the use of positional embedding in \gate can consistently achieve better accuracy compared to the design without positional embedding.

\subsection{Distillation/Training}
\label{subsec:training}

\begin{figure}[h]
    \centering
    \includegraphics[width=\linewidth]{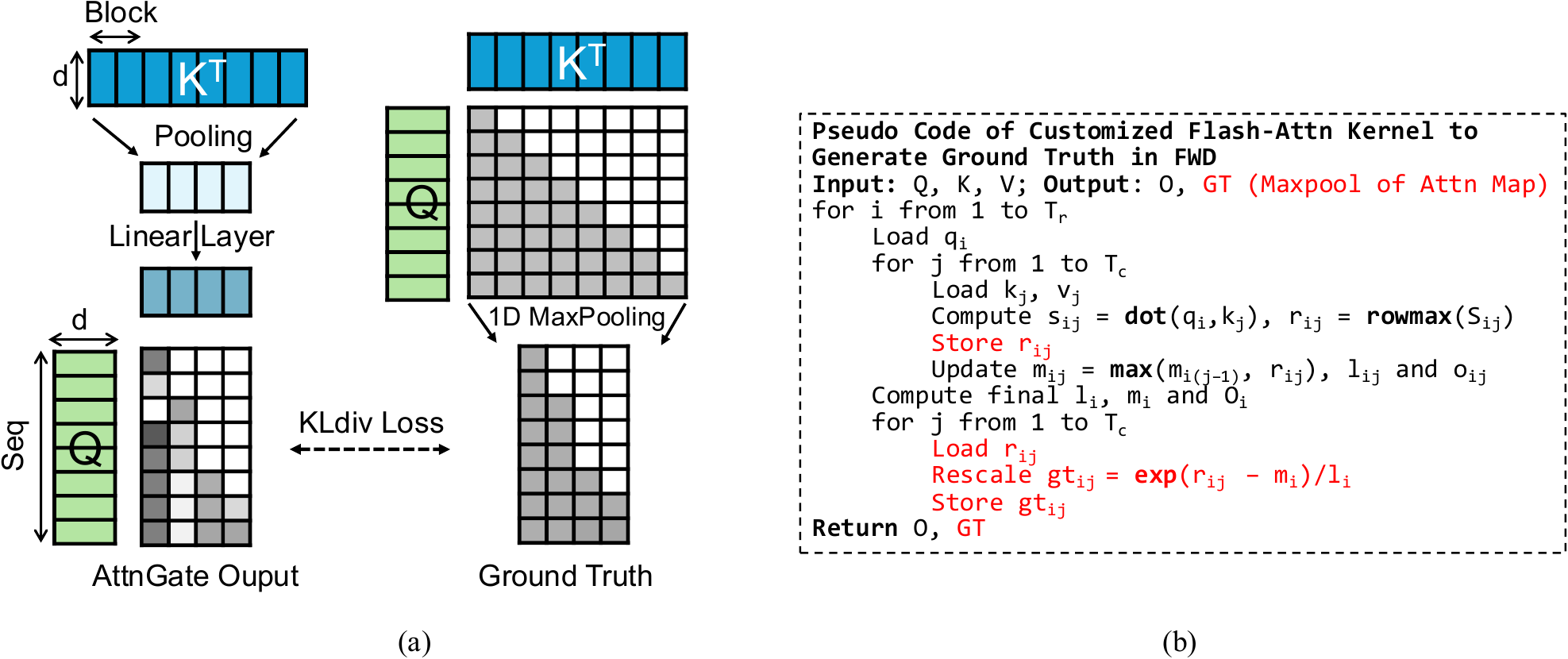}
    \caption{Training Diagram and Training Kernel of \seerR. (a) Self-distillation training of \gate in \seerR. It uses 1D maxpooled attention scores from original model as ground truth to train \gate. Query head reduction is not plotted in the diagram for simplicity. (2) Pseudo code of attention forward kernel for training that directly generates ground truth and attention output.}
    \label{fig:training}
\end{figure}

Previous \seer introduces \gate distillation method using the ground truth generate by LLM itself in the prefilling phase. The training process is efficient as only the \gate are trained. In \seerR, we extend this method to the decoding scenario by slightly changing the form of the ground truth. \Figref{fig:training} shows the overall diagram of the training process.

\paragraph{Ground Truth}
To train \gate for the auto-regressive decoding process, we need to adapt the ground truth generation method. Instead of performing 2D maxpooling of attention map in the prefill case, we only do column-wise 1D maxpooling shown in \Figref{fig:training}a. This corresponds to the decoding \gate that does not compress in sequence dimension. Moreover, to accommodate the shared sparsity in GQA, the column-pooled attention map is further maxpooled within each query heads subgroup, resulting in a ground truth with key-value heads. Finally, the ground truth is normalized to summation 1. We then use the Kullback-Leibler divergence loss~\citep{kl} to train \gate in the distillation process.

\paragraph{Efficiently Obtaining Ground Truth during Training}
Explicitly calculating the full attention map softmax($\mathbf{QK^T/\sqrt{d}}$) and then perform the block-level pooling can cost huge GPU memory due to the quadratic complexity. In \seerR, we also provide an efficient modification of FlashAttention-2~\cite{flash2} kernel that directly generates the ground truth along with the attention output. This kernel largely reuses the intermediate results (e.g. block-level rowmax) in Flash-Attention and thus increases the efficiency of the distillation process. The pseudo code is shown in \Figref{fig:training}b.

\section{Inference of \seerR}
\begin{figure}[h]
    \centering
    \includegraphics[width=0.8\linewidth]{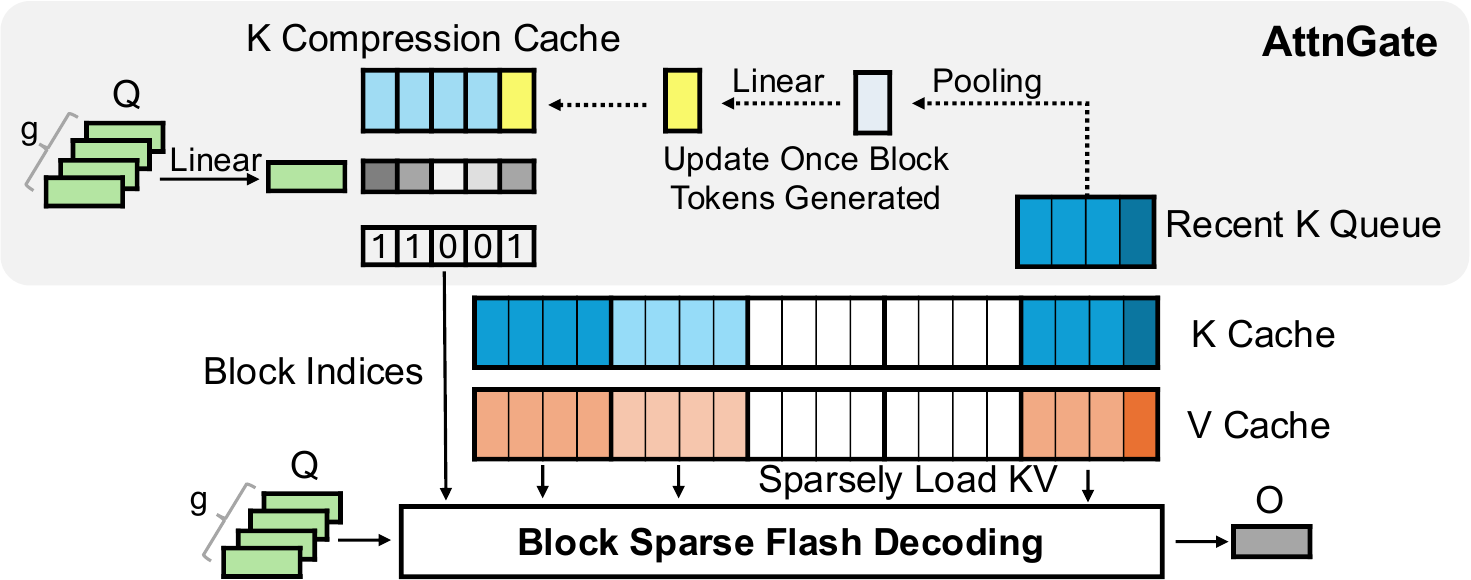}
    \caption{Inference Diagram of \seerR. During inference, a K Compression Cache is used to cache the compressed key representation in \gate to speedup sparse block prediction. This K Compression Cache only updates once per block number of tokens is generated (block=4 in the plots for illustration). As a result, the last block of sequence is always selected to compensate when the compression cache has not been updated yet. $g$ is the group size of GQA.}
    \label{fig:inference}
\end{figure}

\subsection{Sparsify Methods: Token Budget vs Threshold}
During training, the \gate output $\mathbf{S}$ are distilled to mimic the distribution of the block-wise attention maps from the original model in real-valued (floating-point) form. During inference, important key-value blocks can be selectively activated based on the predictions of \gate. In \seerR, we apply two sparsity methods to convert the soft \gate outputs into binary block masks (or block indices).
The first method is the \textit{token budget} approach, which is widely adopted in sparse attention methods. Given a fixed token budget, it is first translated into a block budget by dividing the token budget by the block size. The \gate outputs are then sorted using a Top-k kernel, where k corresponds to the block budget. While this method introduces an additional Top-k operation, it eliminates the need for a softmax operation in \gate.
The second method is the \textit{threshold} approach, which simply selects blocks whose scores exceed a given threshold. The threshold method is more self-adaptive as different heads may automatically infer different sparsity ratios. While these two methods involve different trade-offs between efficiency and accuracy, the token budget approach is better suited for direct comparisons with other methods.

\subsection{K Compression Cache}
Similar to KV cache, in \seerR, we use a \textit{K Compression Cache} to store the compressed representation of K (after pooling plus linear) to speedup \gate prediction. Thus, \gate does not need to recompute $K$ branch for past seen tokens. The update of K Compression Cache is consist of two phases. First, when the sequence length is not the multiplies of block size $b$, the new entry of K Compression Cache may not be accurate. During this time, the last block is always activated to eliminate unnecessary accuracy loss. Second, as long as $b$ number of new tokens are generated, the most recent $b$ tokens will pass through the pooling and linear layer and update the K Compression Cache. In this way, the overhead of \gate can be minimized. 

In practice, \seerR utilizes a relatively large block size $b$, such as 64, which significantly reduces the overhead of the K Compression Cache. Specifically when $b=64$, the additional memory required for the K Compression Cache amounts to only 1/128 (<1\%) of the original KV cache size. This minimal overhead makes it highly efficient. Moreover, it introduces the possibility of offloading the larger KV cache to CPU or other storage. During inference, only the activated blocks need to be retrieved and transferred back to GPU memory on demand. Alternatively, sparse attention computations can even be performed on heterogeneous resources, such as the CPU, further optimizing memory usage and enabling efficient handling of long-context decoding tasks.

\subsection{Block Sparse Flash Decoding Kernel}
\label{subsec:kernel}
To accelerate decoding under block-sparse attention, we design a specialized kernel that extends the FlashAttention decoding pattern to support dynamic block sparsity in the key/value memory. Our kernel adopts the grid scheduling strategy of flash decoding for GQA, using a three-dimensional launch space over (\textit{batch, heads\_kv, num\_split}). This design supports concurrent computation across multiple query groups and key/value shards, maximizing block-level parallelism.

Our block sparse version of the decoding kernel takes the activated block indices from \gate (shape [\textit{batch, heads\_kv, max\_selected\_blocks}]), which encodes the selected key/value blocks for each group of query heads. 
During execution, the kernel only traverses the selected indices and thus skips invalid entries, avoiding unnecessary computation and memory access.
To improve load/compute balancing across Streaming Multiprocessors (SMs), we partition the key/value blocks along the \textit{num\_split} dimension using \textit{max\_selected\_blocks} rather than the total number of blocks. This strategy ensures a more uniform work distribution in the presence of sparsity-induced irregularity.

On H100 GPUs, our kernel leverages the \textit{wgmma} instructions for better Tensor Core usage by padding the number of query head groups to 64. We implement the kernel using TileLang~\cite{tilelang}, which automatically applies computation optimizations like tiling~\cite{roller}, warp specialization and pipelining~\cite{pipethreader}, and memory layout optimizations such as tensorization, rasterization and swizzling~\cite{ladder} based on the target architecture. Additionally, we provide a Triton-based implementation with the same scheduling strategy, allowing for comparative evaluation.

\section{Experiments}

\subsection{Experiments Setup}

\paragraph{Benchmarks, Models, and Baselines} We evaluate \seerR on three math reasoning benchmarks: the American Invitational Mathematics Examination: AIME24, AIME25~\cite{aime}, and MATH-500~\cite{math500}, as well as GPQA-Diamond~\cite{gpqa}. For model evaluation, we select four open-source pre-trained language models with strong reasoning capabilities: Qwen3-4B, 8B, 14B~\cite{yang2025qwen3}, and DeepSeek-R1-Distill-Qwen-14B~\cite{r1}. All models are based on the standard Transformer architecture with Grouped Query Attention (GQA). We compare \seerR against standard full attention and Quest~\cite{quest}. Quest is a training-free sparse attention algorithm applied during decoding, employing a query-aware key-value (KV) cache selection strategy. Specifically, Quest estimates the upper bound of attention scores within each KV block (or “page”) to select the most relevant blocks. By default, Quest uses a block size of 16, and keeps the first two layers fully dense to minimize error. In Section~\ref{subsec:results}, we set the block size to 64 for both Quest and \seerR, and apply sparse attention to all layers to enable a direct comparison. We also conduct ablation studies to analyze the impact of varying block sizes (Section~\ref{subsec: ablalation_blocksize}) and incorporating hybrid dense layers (Section~\ref{subsec: ablalation_hybrid}). Note that \seerR enables shared sparsity selection within each GQA group, whereas Quest does not.
Across all experiments, we set the max output length to 32,768 tokens. While Qwen3 series of models extends this length to 38,912 for AIME24 and AIME25 in their official report, we fix this output length to ensure consistency and fair comparison across all settings.
For \seerR and the full attention baseline, we report average pass@1 accuracy over 64 samples for AIME24 and AIME25, 8 samples for MATH-500, and 16 samples for GPQA-Diamond.

\paragraph{Training Setup for \seerR} To distill \gate, we use the OpenR1-MATH-220k~\cite{openr1} dataset for training. Importantly, only the \gate is trained, and the original model weights remain unchanged. Inputs are packed into sequences of up to 32k tokens with our variable-length Flash-Attention training kernel that also generates ground truth (Section~\ref{subsec:training}). Training is performed with a global batch size of 16 for 800 steps on AMD MI300x GPUs, utilizing DeepSpeed ZeRO-2 optimization. We use AdamW optimizer and a learning rate of 1e-3 with cosine decay schedule.

\subsection{Oracle Sparse Accuracy: How Sparse is Attention in Reasoning Models?} \label{subsec:oracle}
\begin{figure}[h]
    \centering
    \includegraphics[width=1\linewidth]{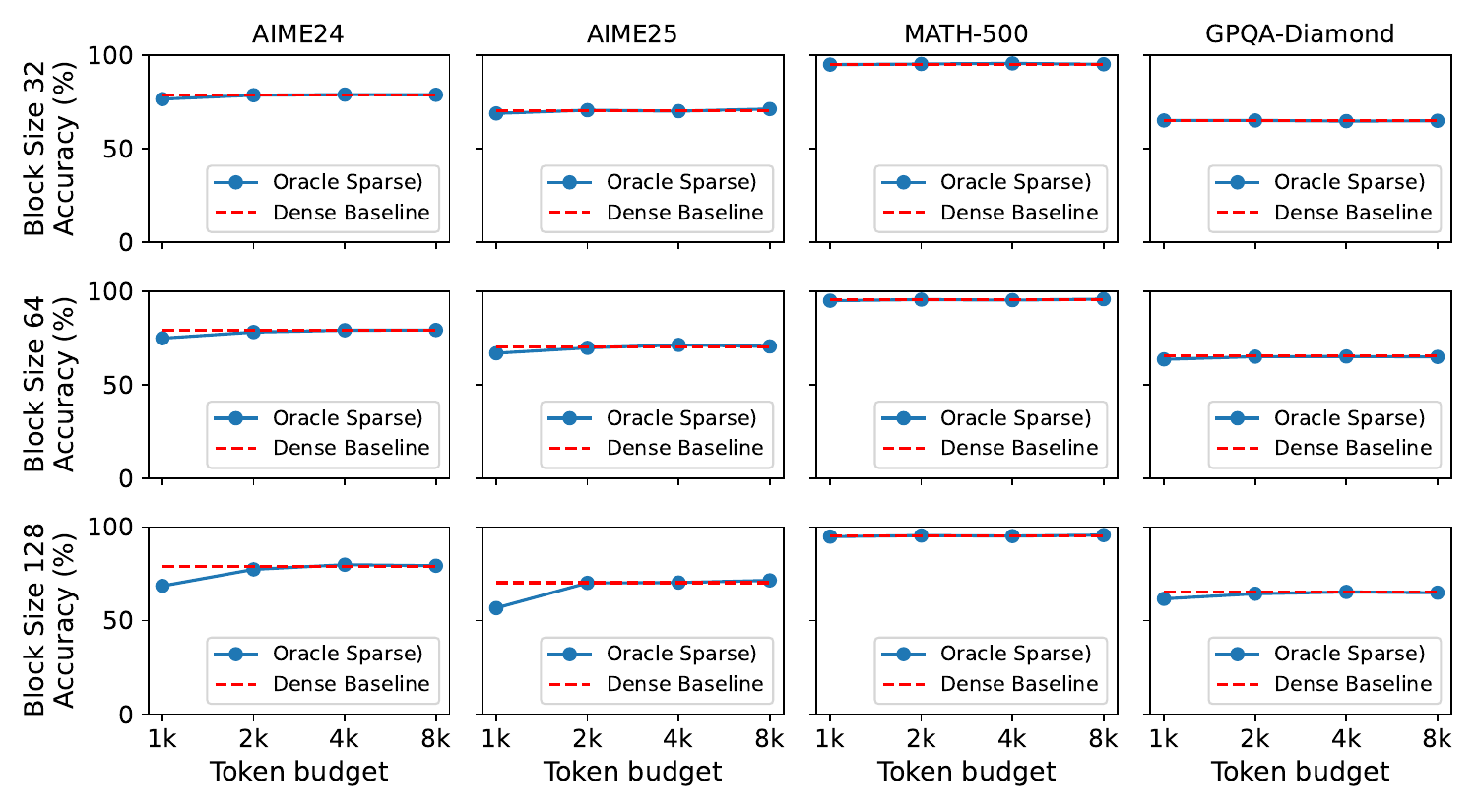}
    \caption{Oracle Sparse Results of Qwen3-14B with block size 32, 64, 128. }
    \label{fig:oracle sparse}
\end{figure}

In the first experiment, we aim to answer the question: \textit{How sparse is attention in reasoning models?} To investigate this, we employ \textit{oracle block sparse selection}, which utilizes the ground truth in \seerR training to select sparse key-value (KV) blocks. While this approach basically means compute attention twice and does not provide any speedup, it allows us to evaluate the accuracy upper bound achievable by \seerR under ideal sparse selection.

We evaluate Qwen3-14B with three different sparse block sizes: 32, 64, and 128. The token budgets range from 1k to 8k. As shown in \Figref{fig:oracle sparse}, using oracle sparsity achieves lossless performance on all tasks when the token budgets reach
2k. For the more challenging AIME24 and AIME25 tasks, some accuracy degradation is observed with 1k token budget, particularly with the largest block size (128). However, this degradation is negligible when using block sizes of 32 or 64. These results indicate that attention sparsity exists in the reasoning process. Based on this, we select a block size of 64 as the default for \seerR.


\subsection{Results of \seerR and Quest} \label{subsec:results}
\begin{figure}[h]
    \centering
    \includegraphics[width=1\linewidth]{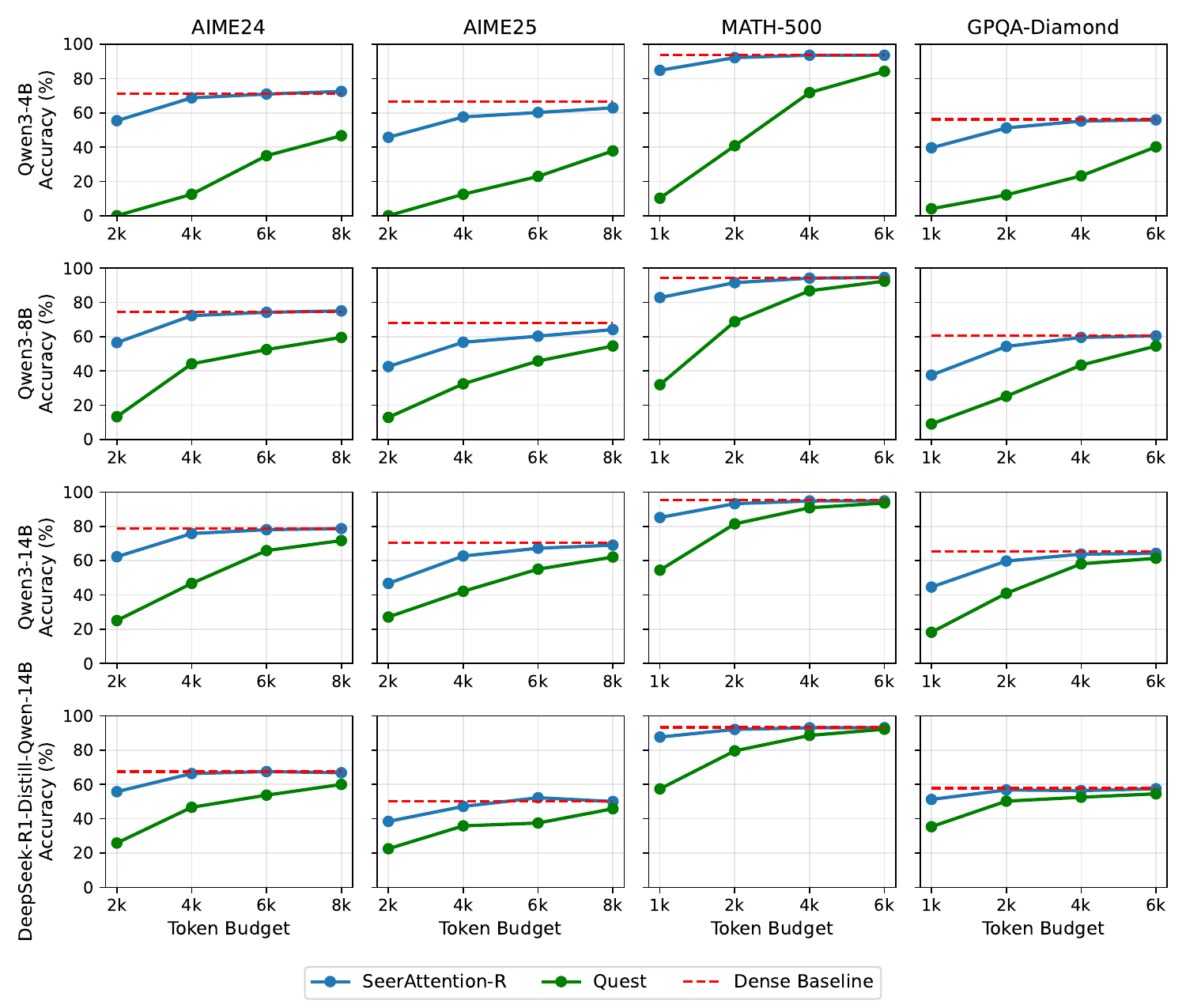}
    \caption{Accuracy Results of Full Attention, \seerR, and Quest. The Quest sparse configuration is set to be the same as \seerR for fair comparison, which uses a block size of 64 and sparse attention in all layers.}
    \label{fig:main results}
\end{figure}

\Figref{fig:main results} shows the results of all the models and benchmarks of Full Attention baseline, \seerR and Quest. As mentioned above, we modify the configuration of Quest to be the same as \seerR (block size 64 and using sparse attention in all layers). We use token budgets from 2k, 4k, 6k, and 8k for AIME24 and AIME25, and 1k, 2k, 4k, and 6k for MATH-500 and GPQA-Diamond. This is mainly because the typical averaged reasoning length from different benchmark is not the same. For the more challenging AIME24 and AIME25, the averaged generated lengths of these models are around 11k-18k. While for the easier MATH-500 and GPQA, the averaged lengths are reduced to 4k-9k. However, it is critical to note that across all combinations, the maximum generation lengths all reached the 32k token cap, underscoring the consistent demand for efficient long-context processing.

The results show that \seerR achieves consistently better performance compared to Quest.  
This trend holds true across every benchmark and computational budget, underscoring the robustness and effectiveness of \seerR. 
For the AIME24 benchmark, \seerR typically achieves lossless performance with 4k token budget on while the previous oracle sparse only requires 2k. This is within expectation as \seerR is only an approximation of ground truth with much less computation required. However, Quest fails achieve lossless accuracy even using 8k token budgets under identical setting. 
For MATH-500 and GPQA-Diamond, the lossless token budgets reduce to 2k for \seerR while Quest requires around 8k to approach the full attention baseline.  

A key trend observed across the results is the relationship between model scale and tolerance for sparse attention. Larger models, such as the 14B variants, exhibit greater robustness to the information loss inherent in sparsity compared to their 4B and 8B counterparts. This phenomenon is particularly pronounced for Quest, where the accuracy gap at lower budgets shrinks significantly as the model size increases. For \seerR, the effect is also present. The 14B models close the final gap to the dense baseline more easily than smaller models on challenging benchmarks like AIME25. This indicates that as reasoning models continue to scale, the viability of sparse attention methods increases.


In conclusion, the results demonstrate the superiority of \seerR's self-distilled approach over the training-free heuristics of Quest, especially in the challenging large block size configuration. Previous work Lserve~\cite{lserve} also mentions the accuracy degradation of Quest over larger block sizes. They resolve this challenge by introducing Hierarchical Paging, a system approach that uses an additional level of block(page) abstraction called virtual logical page, which decouples the sparsity selection page size and physical page size.
With \seerR, we can possibly simplify the sparse attention system design by using a larger block size. 

\subsection{Kernel Speedup}
\begin{figure}[h]
    \centering
    \includegraphics[width=1\linewidth]{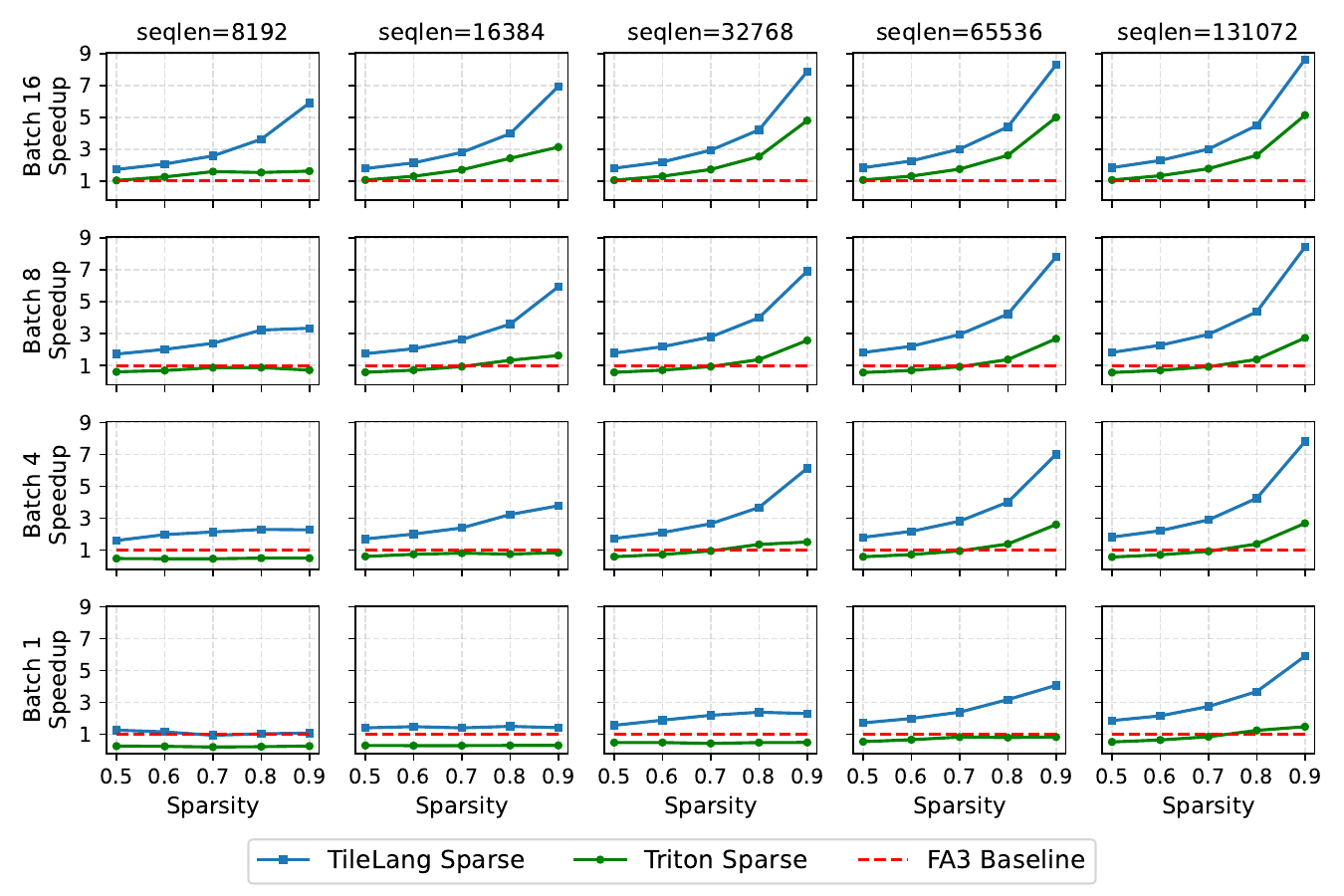}
    \caption{Kernel Speedup of our Block Sparse Flash-Decoding Kernel on H100 GPU. Our TileLang implementation of the kernel achieves higher speedup ratio compared to Triton implementation. For longer sequence length or larger batch size cases, the speedups approach the theoretical upper bound compared to FA3 basline.}
    \label{fig:kernel}
\end{figure}


This section evaluates our customized block sparse flash decoding kernel described in Section~\ref{subsec:kernel}.  We implement the kernel with both TileLang~\cite{tilelang} and Triton and we use FlashAttention-3 (FA3)~\cite{flash3} as baseline. The experiments are run on Nvidia H100 GPU with different input sequence lengths (8k to 128k), batch sizes(1 to 16) and sparsity ratios (0.5 to 0.9). In terms of GQA setting, we use a setting of 64 attention heads with 8 key-value heads, and head dimension 128. 

\Figref{fig:kernel} presents the detailed results. Each subplot corresponds to a specific combination of input sequence length (seqlen) and batch size (bs), with the x-axis showing different sparsity ratios and the y-axis indicating speedup. The TileLang implementation consistently outperforms the FA3 baseline and achieves greater speedup than the Triton implementation. In general, the sparse kernel delivers higher speedup when the input sequence length is longer or the batch size is larger. This is expected, as the decoding kernel is primarily I/O-bound. When the KV cache size is sufficient to saturate the bandwidth, such as when bs=16 and seqlen $\geq$
32k, our sparse kernel achieves near-theoretical speedup (up to 
$9\times$ at 0.9 sparsity). Even for moderate KV cache sizes, e.g. bs=4 and seqlen=32k, the kernel demonstrates significant speedup (up to $6\times$ at 0.9 sparsity).

\section{Ablation Studies}

\subsection{Block Size for Sparse Attention} \label{subsec: ablalation_blocksize}
\begin{figure}[h]
    \centering
    \includegraphics[width=1\linewidth]{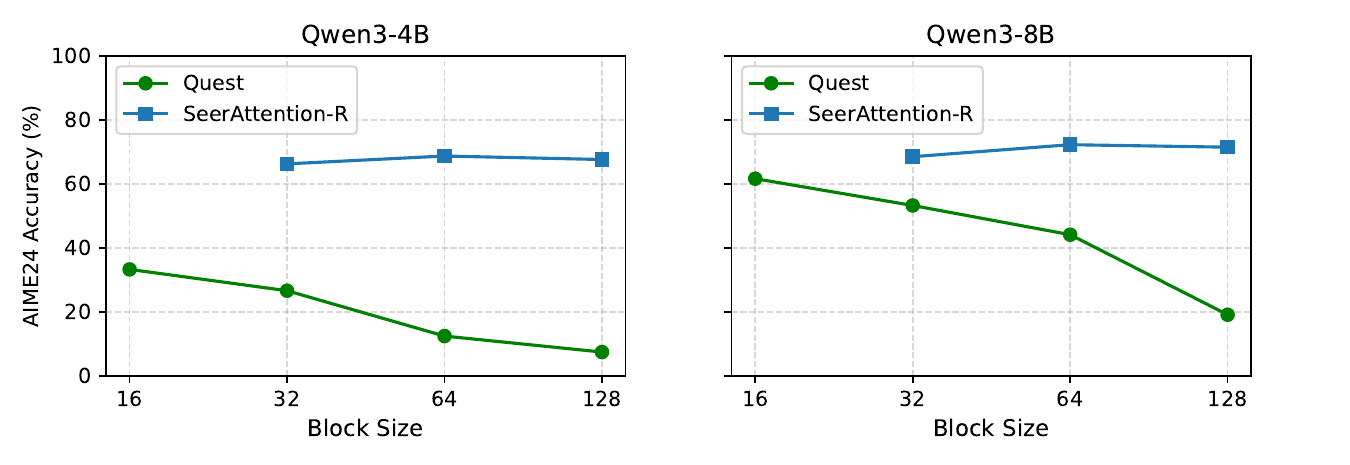}
    \caption{AIME24 results using different block sizes with 4k token budget. \seerR achieves almost consistent performances on different block sizes. However, Quest gets lower accuracy when block size gets larger. Note that in this experiment, \seerR enables shared sparsity selection within each GQA group, whereas Quest does not.
    }
    \label{fig:block size}
\end{figure}



The token block size for sparse attention is a critical factor that affects overall system performance. If the block size is too small, it incurs significant overhead in sparse block prediction, including increased computational cost and larger metadata requirements such as compression caches and block indices.
While a larger block size can also potentially improve the utilization of GPUs. 

\Figref{fig:block size} presents AIME24 results on the Qwen3-4B and Qwen3-8B models across block sizes ranging from 16 to 128. By default, Quest uses a block size of 16. The results indicate that Quest’s performance decreases as the block size increases. However, \seerR achieves consistent accurate sparse block selection at different block sizes. Remarkably, this robustness lies under the assumption of the additional mask sharing in the GQA group dimension. 
We excluded a block size of 16 from our experiments due to its inefficiency during both training and inference. It often leads to out-of-memory errors because of the large intermediate attention maps generated during training.

\subsection{Hybrid Dense Attention in the First Two Layers} \label{subsec: ablalation_hybrid}

\begin{figure}[h]
    \centering
    \includegraphics[width=1\linewidth]{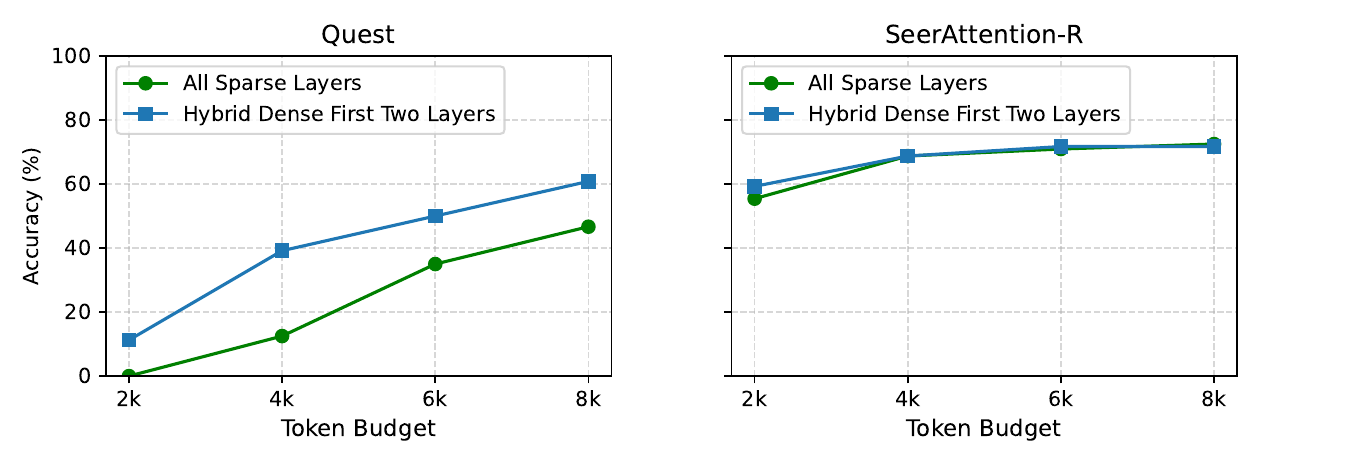}
    \caption{AIME24 results of whether using dense attention in first two layers (Qwen3-4B). }
    \label{fig:hybrid}
\end{figure}

Some post-training sparse attention algorithms employ hybrid dense attention in certain layers to mitigate accuracy loss. By default, Quest applies dense attention in its first two layers. However, for a fair comparison, we evaluate both Quest and \seerR using purely sparse attention across all layers in previous evaluation. This approach allows us to isolate and analyze the effects of sparse attention without the confounding influence of hybrid attention. 

To further investigate the impact of hybrid dense attention, we conduct an ablation study using the Qwen3-4B model on the AIME24 benchmark with a block size of 64. As shown in \Figref{fig:hybrid}, incorporating hybrid dense attention in Quest yields a significant improvement in accuracy, whereas \seerR only sees marginal benefits. This difference may be due to the already accurate sparse prediction by \seerR in the first two layers, reducing the potential gains from hybridization.

\subsection{Threshold VS Token Budgets}

\begin{figure}[h]
  \centering
  \begin{subfigure}[b]{0.43\textwidth}
    \includegraphics[width=\textwidth]{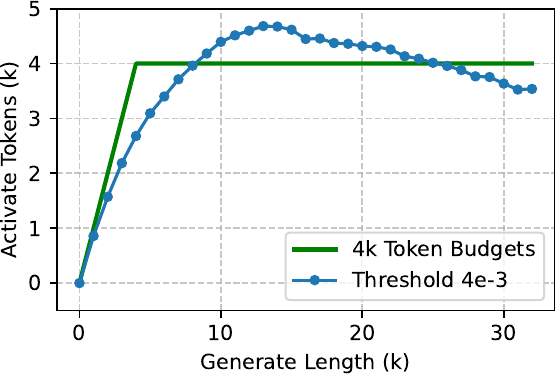}
    \caption{}
    \label{fig:thresholda}
  \end{subfigure}
  \hspace{15pt}
  \begin{subfigure}[b]{0.45\textwidth}
    \includegraphics[width=\textwidth]{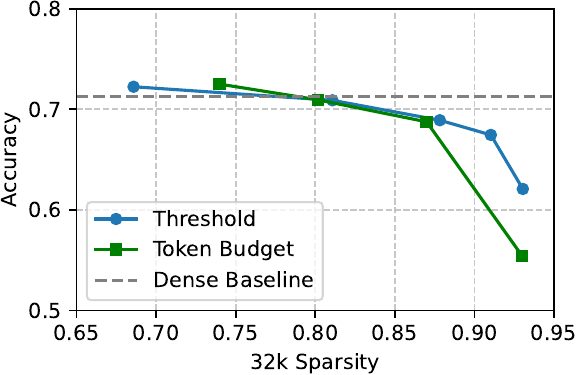}
    \caption{}
    \label{fig:thresholdb}
  \end{subfigure}

  \caption{Threshold vs. Token Budget. Results are obtained using Qwen3-4B models on AIME24 benchmark. (a) Difference of activated tokens distribution of two methods. (b) Sparsity vs Accuracy tradeoff of two methods. Thresholds: 2e-3, 3e-3, 4e-3, 5e-3, 6e-3. Token Budget: 8k, 6k, 4k, 2k. }
  \label{fig:threshold}
\end{figure}

In \seerR, we employ two \gate sparsification strategies, threshold and token budget,  to convert real-valued gate scores into discrete block selections. The token budget method offers an straightforward way to align sparsity and compare with different methods. However, the threshold method is extremely simple to implement and avoids the need of sorting.
\Figref{fig:thresholda} illustrates the distribution of activated tokens across varying sequence lengths using a threshold of 4e-3 and a token budget of 4K on the AIME24 benchmark with Qwen3-4B model. The token budget approach results in a strict piecewise linear activation pattern, whereas the threshold method yields a smoother, curved distribution.
\Figref{fig:thresholdb} compares the sparsity–accuracy trade-offs of the two methods. The threshold method shows slightly better accuracy in high sparsity region.

\subsection{Impact of Sparse Attention on Generate Length}

\begin{table}[h]
\centering
\caption{Qwen3-8B AIME24 Accuracy vs. Reasoning Length.  }
\label{tab:genlen}
\begin{tabular}{cccccc}
\hline
                      &                                     & \multicolumn{4}{c}{Token Budgets} \\
                      &                                     & 2k     & 4k     & 6k     & 8k     \\ \hline
\multicolumn{1}{c|}{\multirow{2}{*}{Quest}}           & \multicolumn{1}{c|}{Accuracy} & 13.3 & 44.2 & 52.5 & 59.6 \\
\multicolumn{1}{c|}{} & \multicolumn{1}{c|}{Gen. Length(k)} & 30.0   & 22.9   & 19.6   & 17.2   \\ \hline
\multicolumn{1}{c|}{\multirow{2}{*}{SeerAttention-R}} & \multicolumn{1}{c|}{Accuracy} & 56.6 & 72.3 & 74.2 & 75.1 \\
\multicolumn{1}{c|}{} & \multicolumn{1}{c|}{Gen. Length(k)} & 19.8   & 16.3   & 15.3   & 15.1   \\ \hline
\end{tabular}
\end{table}

We observed that using inaccurate sparse attention (too small budget or low recall) can increase output token lengths in reasoning tasks. \tabref{tab:genlen} shows the AIME accuracy and reasoning length using Qwen3-8B model. The baseline accuracy of full attention and the generated length are 74.5 and 15.1 k, respectively. We can see that Quest, and \seerR with 2k budget cases, all incur much longer reasoning paths compared to full attention.
A similar phenomenon has been reported in quantization~\cite{rlquant}, where inaccurate quantization algorithms lead to longer reasoning paths. We believe this effect is universal across different post-training efficiency optimizations of reasoning model, as such methods can introduce errors that accumulate over the long reasoning chains. These additional reasoning steps potentially undermine the original goal of improving efficiency.
Therefore, an accurate sparse attention selection algorithm is crucial to mitigate this effect. Another promising approach to eliminate the accumulated errors is to use Rectified Sparse Attention~\cite{resa}, which periodically performs dense rectification of the KV cache.
 
\subsection{Training Budget}

\begin{table}[h]
\centering
\caption{Training Budgets}
\label{tab:training_budget}
\begin{tabular}{c|ccc}
\hline
\textbf{Training Tokens} & \multicolumn{3}{c}{\textbf{GPU Hours}} \\ \hline
\multirow{2}{*}{0.4B}    & Qwen3-4B    & Qwen3-8B   & Qwen3-14B   \\
                         & 10.9        & 12.2       & 18.6        \\ \hline
\end{tabular}
\end{table}

As a lightweight distillation process where only the \gate parameters are trained, \seerR is also highly efficient in terms of training. In our experiments, we set the global batch size to 16 and trained for just 800 steps, utilizing DeepSpeed Stage 2 optimization on MI300x GPUs. Each data batch is packed to a sequence length of 32k with our custom variable-length FlashAttention forward kernel, as described in Section \ref{subsec:training}. \tabref{tab:training_budget} summarizes the GPU hours required for training models of various sizes. Notably, distilling an 8B model requires only 12 GPU hours, demonstrating the efficiency of our approach.
Increasing the quantity, quality, and diversity of training data may lead to further improvements.

\section{Limitation and Future Work}

\subsection{End-to-end Speedup}
The current focus of this work is on the accuracy of sparse decoding and its kernel-level speedup, while leaving end-to-end system support and optimization for future work. Achieving significant end-to-end speedup will require integration with state-of-the-art inference frameworks such as vllm~\cite{vllm}, sglang~\cite{sglang} and Lserve~\cite{lserve}, along with additional support for sparse kernels with PagedAttention. In addition, \seerR can possibly be combined with KV cache offloading technique similar to previous works~\cite{xiao2024infllm,liu2024retrievalattention, magicpig, hao2025omnikv} to save GPU memory and only keep the K Compression Cache with \gate to dynamically control computation/communication.

\subsection{Adaptive Sparsity Ratio}
Determining the optimal sparsity ratio for attention is a non-trivial challenge.
It involves a fundamental trade-off between accuracy and efficiency, and depends on multiple dynamic factors including the input length, task difficulty, and reasoning length.
In general, easier tasks often allow for greater sparsity. However, for reasoning models, more difficult tasks tend to require longer reasoning steps, where attention computation only becomes a bottleneck for longer sequences. This apparent contradiction highlights the need for (1) a sufficiently precise sparsity selection algorithm and (2) automatic adaptation of sparsity ratios based on task complexity. One promising solution is to use Top-p (Nucleus sampling) in sparsity selection, which has previously been explored in Twilight~\cite{twilight_topp} and MagicPIG~\cite{magicpig}. Specifically, Twilight adopts a binary search algorithm to find the optimal threshold that satisfies Top-p.

\subsection{Unify Sparse Prefill and Decoding}

Supporting efficient sparse attention in both prefill and decoding is crucial for end-to-end acceleration in tasks such as autonomous agents and deep research, which involve repeated cycles of long-context prefill and decoding during multi-step reasoning.
However, jointly enabling sparse prefill and decoding, while simultaneously preserving high accuracy and high efficiency, remains an important and active research challenge.
Currently, \seer and \seerR are trained separately, each with its own \gate design. 

From an efficiency perspective, prefill benefits from high query-level parallelism, while auto-regressive decoding lacks such parallelism. Prior work like NSA~\cite{nsa} chooses to discard query-level parallelism and only utilize the group dimension in GQA to enable identical sparse attention scheme in both prefill and decoding. This approach typically requires a larger group size in GQA to achieve better efficiency. 

Another promising direction is to integrate techniques such as multi-token prediction~\cite{gloeckle2024better, liu2024deepseek} or speculative decoding~\cite{leviathan2023fast} into the sparse attention framework. These methods naturally introduce opportunities for query-level parallelism during decoding. While enhancing parallelism and efficiency, they offer a path toward unifying the gating mechanism across prefill and decoding, potentially eliminating the need for separate gating adapters as in \seer and \seerR.
\section{Related Works}

\subsection{Training-free vs. Training-based Sparse Attention}
Sparsity in attention mechanisms has been widely observed in modern neural network architectures. To achieve significant computational speedup, it is essential to identify or predict the locations of non-zero (i.e., important) attention outputs prior to performing the actual attention operation. Research on sparse location identification has generally followed two main directions: training-free (pre-defined or heuristic) algorithms and training-based methods. 
Training-free approaches typically employ static patterns~\cite{streamingllm, moa, duo} or heuristic-based algorithms for sparse location identification~\cite{h2o, quest, minference, lserve, flexprefill, magicpig, liu2024retrievalattention, li2024scbench, doublesparse, raas, zhang2025spargeattn, xu2025xattention, retroinfer}. These methods often rely on human observation or prior knowledge about sparsity, such as fixed patterns or specific head characteristics. 

On the other hand, a different line of research focuses on training-based sparse attention methods to minimize the accuracy degradation from sparse attention. Early work in this area involved integrating various sparse attention patterns (e.g., local, global, and block attention) directly into the model architecture to reduce computational complexity~\cite{sparsetransformer, longformer, zaheer2020big}. More recently, methods such as NSA~\cite{nsa}, MoBA~\cite{moba}, ACP~\cite{acp}, MiniCPM4~\cite{minicpm4} have introduced dynamic sparse attention modules that are trained during LLM pre-training. These native sparse pre-trained models are able to perform sparse attention without incurring additional accuracy loss.
Yet, SeerAttention offers a post-training, trainable approach in the middle ground. It learns the original sparsity patterns without modifying or fine-tuning the original model weights, thereby providing a flexible and efficient solution.

\subsection{KV Cache Compression: A Sparse Attention Perspective}
Optimizing the key-value (KV) cache is a crucial component for efficient LLM inference, as reducing the KV cache size directly translates to savings in I/O bandwidth and GPU memory consumption. Some of the methods involve permanent eviction of tokens in KV cache~\cite{fastgen, snapkv, attentiongatekv, h2o, liu2023scissorhands, adnan2024keyformer, chen2024sepllm, behnam2025rocketkv}, where tokens are chosen using given policies for removal. While this approach can significantly reduce memory usage, it poses potential risks of accuracy loss, as it is difficult to predict which tokens will be important for future generations. On the other hand, dynamic selection without permanent eviction~\cite{quest, pqcache, squeezedattention, retroinfer, liu2024retrievalattention, raas, r-kv, hao2025omnikv, metaclusterkv} allows for tokens to be dynamically selected for attention at each step without irreversibly discarding them.

\subsection{Other Efficient Attention Algorithms}

Apart from sparse attention, efforts have also been devoted to developing efficient attention algorithms. Variations of Multi-head Attention~\cite{attention}, such as Grouped-Query Attention~\cite{gqa}, Multi-Query Attention~\cite{mqa}, Multi-head Latent Attention~\cite{deepseekv2}, as well as recent Grouped-Tied Attention and Grouped Latent Attention~\cite{tridao_efficient_attention}, have been proposed to balance KV cache size, arithmetic intensity and model quality.  In addition to modifying attention within a layer, approaches like YOCO~\cite{yoco} and CLA~\cite{cla} explore cross-layer KV cache sharing to reduce KV cache sizes.  

Linear attention represents another important research direction~\cite{linearattention, retnet, beck2024xlstm, deltanet, gu2023mamba, mamba2, peng2023rwkv, gla}. These models facilitate efficient parallel training through chunk-wise parallelization and ensure constant memory usage during inference via recurrent computation. 
However, current pure linear attention models still struggle in some challenging long-context tasks, like retrieval or reasoning.
In practice, hybrid architectures that combine linear attention with standard full attention have shown performance comparable to full attention models~\cite{dong2024hymba, li2025minimax}.


\section{Conclusion}

This paper introduces \seerR, a lightweight and flexible sparse attention framework that accelerates long decoding in reasoning models. 
Functioning as a plug-in gating, \seerR integrates seamlessly into existing pretrained models without modifying the original parameters, requiring only a lightweight training phase for its new gating parameters.
\seerR maintains near-lossless reasoning accuracy in a post-training setting, even with coarse-grained attention block sizes.
The highly optimized sparse decoding kernel using TileLang achieves a near-theoretical speedup at high sparsity ratios.

\bibliographystyle{plainnat}
\bibliography{example_paper}

\begin{thebibliography}{79}
\providecommand{\natexlab}[1]{#1}
\providecommand{\url}[1]{\texttt{#1}}
\expandafter\ifx\csname urlstyle\endcsname\relax
  \providecommand{\doi}[1]{doi: #1}\else
  \providecommand{\doi}{doi: \begingroup \urlstyle{rm}\Url}\fi

\bibitem[til()]{tilelang}
{TileLang}.
\newblock URL \url{https://github.com/tile-ai/tilelang}.

\bibitem[Adnan et~al.(2024)Adnan, Arunkumar, Jain, Nair, Soloveychik, and Kamath]{adnan2024keyformer}
Muhammad Adnan, Akhil Arunkumar, Gaurav Jain, Prashant~J Nair, Ilya Soloveychik, and Purushotham Kamath.
\newblock Keyformer: Kv cache reduction through key tokens selection for efficient generative inference.
\newblock \emph{Proceedings of Machine Learning and Systems}, 6:\penalty0 114--127, 2024.

\bibitem[Ainslie et~al.(2023)Ainslie, Lee-Thorp, De~Jong, Zemlyanskiy, Lebr{\'o}n, and Sanghai]{gqa}
Joshua Ainslie, James Lee-Thorp, Michiel De~Jong, Yury Zemlyanskiy, Federico Lebr{\'o}n, and Sumit Sanghai.
\newblock Gqa: Training generalized multi-query transformer models from multi-head checkpoints.
\newblock \emph{arXiv preprint arXiv:2305.13245}, 2023.

\bibitem[Beck et~al.(2024)Beck, P{\"o}ppel, Spanring, Auer, Prudnikova, Kopp, Klambauer, Brandstetter, and Hochreiter]{beck2024xlstm}
Maximilian Beck, Korbinian P{\"o}ppel, Markus Spanring, Andreas Auer, Oleksandra Prudnikova, Michael Kopp, G{\"u}nter Klambauer, Johannes Brandstetter, and Sepp Hochreiter.
\newblock xlstm: Extended long short-term memory.
\newblock \emph{arXiv preprint arXiv:2405.04517}, 2024.

\bibitem[Behnam et~al.(2025)Behnam, Fu, Zhao, Tsai, Yu, and Tumanov]{behnam2025rocketkv}
Payman Behnam, Yaosheng Fu, Ritchie Zhao, Po-An Tsai, Zhiding Yu, and Alexey Tumanov.
\newblock Rocketkv: Accelerating long-context llm inference via two-stage kv cache compression.
\newblock \emph{arXiv preprint arXiv:2502.14051}, 2025.

\bibitem[Beltagy et~al.(2020)Beltagy, Peters, and Cohan]{longformer}
Iz~Beltagy, Matthew~E Peters, and Arman Cohan.
\newblock Longformer: The long-document transformer.
\newblock \emph{arXiv preprint arXiv:2004.05150}, 2020.

\bibitem[Brandon et~al.(2024)Brandon, Mishra, Nrusimha, Panda, and Ragan-Kelley]{cla}
William Brandon, Mayank Mishra, Aniruddha Nrusimha, Rameswar Panda, and Jonathan Ragan-Kelley.
\newblock Reducing transformer key-value cache size with cross-layer attention.
\newblock In \emph{The Thirty-eighth Annual Conference on Neural Information Processing Systems}, 2024.

\bibitem[Cai et~al.(2025)Cai, Xiao, Sun, Luo, Zhang, Wan, Li, Zhou, Chang, Gu, Dong, Anandkumar, Asi, and Hu]{r-kv}
Zefan Cai, Wen Xiao, Hanshi Sun, Cheng Luo, Yikai Zhang, Ke~Wan, Yucheng Li, Yeyang Zhou, Li-Wen Chang, Jiuxiang Gu, Zhen Dong, Anima Anandkumar, Abedelkadir Asi, and Junjie Hu.
\newblock R-kv: Redundancy-aware kv cache compression for training-free reasoning models acceleration.
\newblock \emph{arXiv preprint arXiv:2505.24133}, 2025.

\bibitem[Chen et~al.(2024{\natexlab{a}})Chen, Shi, Li, Gao, Ren, Chen, Jiang, Li, Liu, and Huang]{chen2024sepllm}
Guoxuan Chen, Han Shi, Jiawei Li, Yihang Gao, Xiaozhe Ren, Yimeng Chen, Xin Jiang, Zhenguo Li, Weiyang Liu, and Chao Huang.
\newblock Sepllm: Accelerate large language models by compressing one segment into one separator.
\newblock \emph{arXiv preprint arXiv:2412.12094}, 2024{\natexlab{a}}.

\bibitem[Chen et~al.(2025)Chen, Zhang, Lu, Zhang, Zhang, Luo, Liu, Jiang, Chen, Liu, Ding, Yan, Jiang, Chen, Zhang, Yang, Yang, and Yang]{retroinfer}
Yaoqi Chen, Jinkai Zhang, Baotong Lu, Qianxi Zhang, Chengruidong Zhang, Jingjia Luo, Di~Liu, Huiqiang Jiang, Qi~Chen, Jing Liu, Bailu Ding, Xiao Yan, Jiawei Jiang, Chen Chen, Mingxing Zhang, Yuqing Yang, Fan Yang, and Mao Yang.
\newblock Retroinfer: A vector-storage approach for scalable long-context llm inference, 2025.
\newblock URL \url{https://arxiv.org/abs/2505.02922}.

\bibitem[Chen et~al.(2024{\natexlab{b}})Chen, Sadhukhan, Ye, Zhou, Zhang, Nolte, Tian, Douze, Bottou, Jia, et~al.]{magicpig}
Zhuoming Chen, Ranajoy Sadhukhan, Zihao Ye, Yang Zhou, Jianyu Zhang, Niklas Nolte, Yuandong Tian, Matthijs Douze, Leon Bottou, Zhihao Jia, et~al.
\newblock Magicpig: Lsh sampling for efficient llm generation.
\newblock \emph{arXiv preprint arXiv:2410.16179}, 2024{\natexlab{b}}.

\bibitem[Cheng et~al.(2025)Cheng, Wang, Shi, Xia, Ma, Xue, Wang, Mo, Chen, Yang, Yang, and Yang]{pipethreader}
Yu~Cheng, Lei Wang, Yining Shi, Yuqing Xia, Lingxiao Ma, Jilong Xue, Yang Wang, Zhiwen Mo, Feiyang Chen, Fan Yang, Mao Yang, and Zhi Yang.
\newblock {PipeThreader}: Software-defined pipelining for efficient dnn execution.
\newblock In \emph{19th USENIX Symposium on Operating Systems Design and Implementation (OSDI 25)}, 2025.
\newblock URL \url{https://www.usenix.org/conference/osdi25/presentation/cheng}.

\bibitem[Child et~al.(2019)Child, Gray, Radford, and Sutskever]{sparsetransformer}
Rewon Child, Scott Gray, Alec Radford, and Ilya Sutskever.
\newblock Generating long sequences with sparse transformers.
\newblock \emph{arXiv preprint arXiv:1904.10509}, 2019.

\bibitem[Dao(2023)]{flash2}
Tri Dao.
\newblock Flashattention-2: Faster attention with better parallelism and work partitioning.
\newblock 2023.
\newblock URL \url{https://arxiv.org/abs/2307.08691}.

\bibitem[Dao and Gu(2024)]{mamba2}
Tri Dao and Albert Gu.
\newblock Transformers are ssms: Generalized models and efficient algorithms through structured state space duality.
\newblock \emph{arXiv preprint arXiv:2405.21060}, 2024.

\bibitem[Dong et~al.(2024)Dong, Fu, Diao, Byeon, Chen, Mahabaleshwarkar, Liu, Van~Keirsbilck, Chen, Suhara, et~al.]{dong2024hymba}
Xin Dong, Yonggan Fu, Shizhe Diao, Wonmin Byeon, Zijia Chen, Ameya~Sunil Mahabaleshwarkar, Shih-Yang Liu, Matthijs Van~Keirsbilck, Min-Hung Chen, Yoshi Suhara, et~al.
\newblock Hymba: A hybrid-head architecture for small language models.
\newblock \emph{arXiv preprint arXiv:2411.13676}, 2024.

\bibitem[Face(2025)]{openr1}
Hugging Face.
\newblock Open r1: A fully open reproduction of deepseek-r1, January 2025.
\newblock URL \url{https://github.com/huggingface/open-r1}.

\bibitem[Fu et~al.(2024)Fu, Huang, Ning, Zhang, Chen, Wu, Wang, Huang, Li, Yan, et~al.]{moa}
Tianyu Fu, Haofeng Huang, Xuefei Ning, Genghan Zhang, Boju Chen, Tianqi Wu, Hongyi Wang, Zixiao Huang, Shiyao Li, Shengen Yan, et~al.
\newblock Moa: Mixture of sparse attention for automatic large language model compression.
\newblock \emph{arXiv preprint arXiv:2406.14909}, 2024.

\bibitem[Gao et~al.(2024)Gao, Zeng, Du, Cao, Zhou, Qi, Lai, So, Cao, Yang, et~al.]{seerattn_v1}
Yizhao Gao, Zhichen Zeng, Dayou Du, Shijie Cao, Peiyuan Zhou, Jiaxing Qi, Junjie Lai, Hayden Kwok-Hay So, Ting Cao, Fan Yang, et~al.
\newblock Seerattention: Learning intrinsic sparse attention in your llms.
\newblock \emph{arXiv preprint arXiv:2410.13276}, 2024.

\bibitem[Ge et~al.(2023)Ge, Zhang, Liu, Zhang, Han, and Gao]{fastgen}
Suyu Ge, Yunan Zhang, Liyuan Liu, Minjia Zhang, Jiawei Han, and Jianfeng Gao.
\newblock Model tells you what to discard: Adaptive kv cache compression for llms.
\newblock \emph{arXiv preprint arXiv:2310.01801}, 2023.

\bibitem[Gloeckle et~al.(2024)Gloeckle, Idrissi, Rozi{\`e}re, Lopez-Paz, and Synnaeve]{gloeckle2024better}
Fabian Gloeckle, Badr~Youbi Idrissi, Baptiste Rozi{\`e}re, David Lopez-Paz, and Gabriel Synnaeve.
\newblock Better \& faster large language models via multi-token prediction.
\newblock \emph{arXiv preprint arXiv:2404.19737}, 2024.

\bibitem[Gu and Dao(2023)]{gu2023mamba}
Albert Gu and Tri Dao.
\newblock Mamba: Linear-time sequence modeling with selective state spaces.
\newblock \emph{arXiv preprint arXiv:2312.00752}, 2023.

\bibitem[Guo et~al.(2025)Guo, Yang, Zhang, Song, Zhang, Xu, Zhu, Ma, Wang, Bi, et~al.]{r1}
Daya Guo, Dejian Yang, Haowei Zhang, Junxiao Song, Ruoyu Zhang, Runxin Xu, Qihao Zhu, Shirong Ma, Peiyi Wang, Xiao Bi, et~al.
\newblock Deepseek-r1: Incentivizing reasoning capability in llms via reinforcement learning.
\newblock \emph{arXiv preprint arXiv:2501.12948}, 2025.

\bibitem[Hao et~al.(2025)Hao, Zhu, Wang, Yu, Xin, Zheng, Ren, and Guo]{hao2025omnikv}
Jitai Hao, Yuke Zhu, Tian Wang, Jun Yu, Xin Xin, Bo~Zheng, Zhaochun Ren, and Sheng Guo.
\newblock Omnikv: Dynamic context selection for efficient long-context llms.
\newblock In \emph{The Thirteenth International Conference on Learning Representations}, 2025.

\bibitem[Hendrycks et~al.(2020)Hendrycks, Burns, Basart, Zou, Mazeika, Song, and Steinhardt]{math500}
Dan Hendrycks, Collin Burns, Steven Basart, Andy Zou, Mantas Mazeika, Dawn Song, and Jacob Steinhardt.
\newblock Measuring massive multitask language understanding.
\newblock \emph{arXiv preprint arXiv:2009.03300}, 2020.

\bibitem[Hooper et~al.(2024)Hooper, Kim, Mohammadzadeh, Maheswaran, Paik, Mahoney, Keutzer, and Gholami]{squeezedattention}
Coleman Hooper, Sehoon Kim, Hiva Mohammadzadeh, Monishwaran Maheswaran, June Paik, Michael~W Mahoney, Kurt Keutzer, and Amir Gholami.
\newblock Squeezed attention: Accelerating long context length llm inference.
\newblock \emph{arXiv preprint arXiv:2411.09688}, 2024.

\bibitem[Hu et~al.(2025)Hu, Huang, Wang, Li, Hu, Liu, Chen, Xie, and Shan]{raas}
Junhao Hu, Wenrui Huang, Weidong Wang, Zhenwen Li, Tiancheng Hu, Zhixia Liu, Xusheng Chen, Tao Xie, and Yizhou Shan.
\newblock Efficient long-decoding inference with reasoning-aware attention sparsity.
\newblock \emph{arXiv preprint arXiv:2502.11147}, 2025.

\bibitem[Jaech et~al.(2024)Jaech, Kalai, Lerer, Richardson, El-Kishky, Low, Helyar, Madry, Beutel, Carney, et~al.]{o1}
Aaron Jaech, Adam Kalai, Adam Lerer, Adam Richardson, Ahmed El-Kishky, Aiden Low, Alec Helyar, Aleksander Madry, Alex Beutel, Alex Carney, et~al.
\newblock Openai o1 system card.
\newblock \emph{arXiv preprint arXiv:2412.16720}, 2024.

\bibitem[Jiang et~al.(2024)Jiang, Li, Zhang, Wu, Luo, Ahn, Han, Abdi, Li, Lin, et~al.]{minference}
Huiqiang Jiang, Yucheng Li, Chengruidong Zhang, Qianhui Wu, Xufang Luo, Surin Ahn, Zhenhua Han, Amir~H Abdi, Dongsheng Li, Chin-Yew Lin, et~al.
\newblock Minference 1.0: Accelerating pre-filling for long-context llms via dynamic sparse attention.
\newblock \emph{arXiv preprint arXiv:2407.02490}, 2024.

\bibitem[Joyce(2011)]{kl}
James~M Joyce.
\newblock Kullback-leibler divergence.
\newblock In \emph{International encyclopedia of statistical science}, pages 720--722. Springer, 2011.

\bibitem[Katharopoulos et~al.(2020)Katharopoulos, Vyas, Pappas, and Fleuret]{linearattention}
Angelos Katharopoulos, Apoorv Vyas, Nikolaos Pappas, and Fran{\c{c}}ois Fleuret.
\newblock Transformers are rnns: Fast autoregressive transformers with linear attention.
\newblock In \emph{International conference on machine learning}, pages 5156--5165. PMLR, 2020.

\bibitem[Kwon et~al.(2023)Kwon, Li, Zhuang, Sheng, Zheng, Yu, Gonzalez, Zhang, and Stoica]{vllm}
Woosuk Kwon, Zhuohan Li, Siyuan Zhuang, Ying Sheng, Lianmin Zheng, Cody~Hao Yu, Joseph~E. Gonzalez, Hao Zhang, and Ion Stoica.
\newblock Efficient memory management for large language model serving with pagedattention.
\newblock In \emph{Proceedings of the ACM SIGOPS 29th Symposium on Operating Systems Principles}, 2023.

\bibitem[Lai et~al.(2025)Lai, Lu, Luo, Ma, and Zhou]{flexprefill}
Xunhao Lai, Jianqiao Lu, Yao Luo, Yiyuan Ma, and Xun Zhou.
\newblock Flexprefill: A context-aware sparse attention mechanism for efficient long-sequence inference.
\newblock \emph{arXiv preprint arXiv:2502.20766}, 2025.

\bibitem[Leviathan et~al.(2023)Leviathan, Kalman, and Matias]{leviathan2023fast}
Yaniv Leviathan, Matan Kalman, and Yossi Matias.
\newblock Fast inference from transformers via speculative decoding.
\newblock In \emph{International Conference on Machine Learning}, pages 19274--19286. PMLR, 2023.

\bibitem[Li et~al.(2025)Li, Gong, Yang, Shan, Liu, Zhu, Zhang, Guo, Chen, Li, et~al.]{li2025minimax}
Aonian Li, Bangwei Gong, Bo~Yang, Boji Shan, Chang Liu, Cheng Zhu, Chunhao Zhang, Congchao Guo, Da~Chen, Dong Li, et~al.
\newblock Minimax-01: Scaling foundation models with lightning attention.
\newblock \emph{arXiv preprint arXiv:2501.08313}, 2025.

\bibitem[Li et~al.(2024{\natexlab{a}})Li, Jiang, Wu, Luo, Ahn, Zhang, Abdi, Li, Gao, Yang, et~al.]{li2024scbench}
Yucheng Li, Huiqiang Jiang, Qianhui Wu, Xufang Luo, Surin Ahn, Chengruidong Zhang, Amir~H Abdi, Dongsheng Li, Jianfeng Gao, Yuqing Yang, et~al.
\newblock Scbench: A kv cache-centric analysis of long-context methods.
\newblock \emph{arXiv preprint arXiv:2412.10319}, 2024{\natexlab{a}}.

\bibitem[Li et~al.(2024{\natexlab{b}})Li, Huang, Yang, Venkitesh, Locatelli, Ye, Cai, Lewis, and Chen]{snapkv}
Yuhong Li, Yingbing Huang, Bowen Yang, Bharat Venkitesh, Acyr Locatelli, Hanchen Ye, Tianle Cai, Patrick Lewis, and Deming Chen.
\newblock Snapkv: Llm knows what you are looking for before generation.
\newblock \emph{Advances in Neural Information Processing Systems}, 37:\penalty0 22947--22970, 2024{\natexlab{b}}.

\bibitem[Lin et~al.(2025{\natexlab{a}})Lin, Tang, Yang, Wang, Tang, Tian, Stoica, Han, and Gao]{twilight_topp}
Chaofan Lin, Jiaming Tang, Shuo Yang, Hanshuo Wang, Tian Tang, Boyu Tian, Ion Stoica, Song Han, and Mingyu Gao.
\newblock Twilight: Adaptive attention sparsity with hierarchical top-$ p $ pruning.
\newblock \emph{arXiv preprint arXiv:2502.02770}, 2025{\natexlab{a}}.

\bibitem[Lin et~al.(2025{\natexlab{b}})Lin, Obando-Ceron, He, and Courville]{acp}
Zhixuan Lin, Johan Obando-Ceron, Xu~Owen He, and Aaron Courville.
\newblock Adaptive computation pruning for the forgetting transformer.
\newblock \emph{arXiv preprint arXiv:2504.06949}, 2025{\natexlab{b}}.

\bibitem[Liu et~al.(2024{\natexlab{a}})Liu, Feng, Wang, Wang, Liu, Zhao, Dengr, Ruan, Dai, Guo, et~al.]{deepseekv2}
Aixin Liu, Bei Feng, Bin Wang, Bingxuan Wang, Bo~Liu, Chenggang Zhao, Chengqi Dengr, Chong Ruan, Damai Dai, Daya Guo, et~al.
\newblock Deepseek-v2: A strong, economical, and efficient mixture-of-experts language model.
\newblock \emph{arXiv preprint arXiv:2405.04434}, 2024{\natexlab{a}}.

\bibitem[Liu et~al.(2024{\natexlab{b}})Liu, Feng, Xue, Wang, Wu, Lu, Zhao, Deng, Zhang, Ruan, et~al.]{liu2024deepseek}
Aixin Liu, Bei Feng, Bing Xue, Bingxuan Wang, Bochao Wu, Chengda Lu, Chenggang Zhao, Chengqi Deng, Chenyu Zhang, Chong Ruan, et~al.
\newblock Deepseek-v3 technical report.
\newblock \emph{arXiv preprint arXiv:2412.19437}, 2024{\natexlab{b}}.

\bibitem[Liu et~al.(2024{\natexlab{c}})Liu, Chen, Lu, Jiang, Han, Zhang, Chen, Zhang, Ding, Zhang, et~al.]{liu2024retrievalattention}
Di~Liu, Meng Chen, Baotong Lu, Huiqiang Jiang, Zhenhua Han, Qianxi Zhang, Qi~Chen, Chengruidong Zhang, Bailu Ding, Kai Zhang, et~al.
\newblock Retrievalattention: Accelerating long-context llm inference via vector retrieval.
\newblock \emph{arXiv preprint arXiv:2409.10516}, 2024{\natexlab{c}}.

\bibitem[Liu et~al.(2025)Liu, Sun, Zhang, Bai, Yu, Yu, Yuan, and Hou]{rlquant}
Ruikang Liu, Yuxuan Sun, Manyi Zhang, Haoli Bai, Xianzhi Yu, Tiezheng Yu, Chun Yuan, and Lu~Hou.
\newblock Quantization hurts reasoning? an empirical study on quantized reasoning models.
\newblock \emph{arXiv preprint arXiv:2504.04823}, 2025.

\bibitem[Liu et~al.(2023)Liu, Desai, Liao, Wang, Xie, Xu, Kyrillidis, and Shrivastava]{liu2023scissorhands}
Zichang Liu, Aditya Desai, Fangshuo Liao, Weitao Wang, Victor Xie, Zhaozhuo Xu, Anastasios Kyrillidis, and Anshumali Shrivastava.
\newblock Scissorhands: Exploiting the persistence of importance hypothesis for llm kv cache compression at test time.
\newblock \emph{Advances in Neural Information Processing Systems}, 36:\penalty0 52342--52364, 2023.

\bibitem[Lu et~al.(2025)Lu, Jiang, Liu, Du, Jiang, Hong, Liu, He, Yuan, Wang, Huang, Yuan, Xu, Xu, Lai, Chen, Zheng, Yan, Su, Wu, Zhang, Yang, Zhou, Zhang, and Qiu]{moba}
Enzhe Lu, Zhejun Jiang, Jingyuan Liu, Yulun Du, Tao Jiang, Chao Hong, Shaowei Liu, Weiran He, Enming Yuan, Yuzhi Wang, Zhiqi Huang, Huan Yuan, Suting Xu, Xinran Xu, Guokun Lai, Yanru Chen, Huabin Zheng, Junjie Yan, Jianlin Su, Yuxin Wu, Yutao Zhang, Zhilin Yang, Xinyu Zhou, Mingxing Zhang, and Jiezhong Qiu.
\newblock Moba: Mixture of block attention for long-context llms.
\newblock \emph{arXiv preprint arXiv:2502.13189}, 2025.

\bibitem[Mazar{\'e} et~al.(2025{\natexlab{a}})Mazar{\'e}, Szilvasy, Lomeli, Massa, Murray, J{\'e}gou, and Douze]{metaclusterkv}
Pierre-Emmanuel Mazar{\'e}, Gergely Szilvasy, Maria Lomeli, Francisco Massa, Naila Murray, Herv{\'e} J{\'e}gou, and Matthijs Douze.
\newblock Inference-time sparse attention with asymmetric indexing.
\newblock \emph{arXiv preprint arXiv:2502.08246}, 2025{\natexlab{a}}.

\bibitem[Mazar{\'e} et~al.(2025{\natexlab{b}})Mazar{\'e}, Szilvasy, Lomeli, Massa, Murray, J{\'e}gou, and Douze]{shared_sparse_gqa}
Pierre-Emmanuel Mazar{\'e}, Gergely Szilvasy, Maria Lomeli, Francisco Massa, Naila Murray, Herv{\'e} J{\'e}gou, and Matthijs Douze.
\newblock Inference-time sparse attention with asymmetric indexing.
\newblock \emph{arXiv preprint arXiv:2502.08246}, 2025{\natexlab{b}}.

\bibitem[of~Problem~Solving()]{aime}
Art of~Problem~Solving.
\newblock Aime problems and solutions.
\newblock \url{https://artofproblemsolving.com/wiki/index.php/AIME_Problems_and_Solutions}.

\bibitem[Peng et~al.(2023)Peng, Alcaide, Anthony, Albalak, Arcadinho, Biderman, Cao, Cheng, Chung, Grella, et~al.]{peng2023rwkv}
Bo~Peng, Eric Alcaide, Quentin Anthony, Alon Albalak, Samuel Arcadinho, Stella Biderman, Huanqi Cao, Xin Cheng, Michael Chung, Matteo Grella, et~al.
\newblock Rwkv: Reinventing rnns for the transformer era.
\newblock \emph{arXiv preprint arXiv:2305.13048}, 2023.

\bibitem[Rein et~al.(2024)Rein, Hou, Stickland, Petty, Pang, Dirani, Michael, and Bowman]{gpqa}
David Rein, Betty~Li Hou, Asa~Cooper Stickland, Jackson Petty, Richard~Yuanzhe Pang, Julien Dirani, Julian Michael, and Samuel~R Bowman.
\newblock Gpqa: A graduate-level google-proof q\&a benchmark.
\newblock In \emph{First Conference on Language Modeling}, 2024.

\bibitem[Shah et~al.(2024)Shah, Bikshandi, Zhang, Thakkar, Ramani, and Dao]{flash3}
Jay Shah, Ganesh Bikshandi, Ying Zhang, Vijay Thakkar, Pradeep Ramani, and Tri Dao.
\newblock Flashattention-3: Fast and accurate attention with asynchrony and low-precision.
\newblock \emph{Advances in Neural Information Processing Systems}, 37:\penalty0 68658--68685, 2024.

\bibitem[Shazeer(2019)]{mqa}
Noam Shazeer.
\newblock Fast transformer decoding: One write-head is all you need.
\newblock \emph{arXiv preprint arXiv:1911.02150}, 2019.

\bibitem[Su et~al.(2024)Su, Ahmed, Lu, Pan, Bo, and Liu]{rope}
Jianlin Su, Murtadha Ahmed, Yu~Lu, Shengfeng Pan, Wen Bo, and Yunfeng Liu.
\newblock Roformer: Enhanced transformer with rotary position embedding.
\newblock \emph{Neurocomputing}, 568:\penalty0 127063, 2024.

\bibitem[Sun et~al.(2023)Sun, Dong, Huang, Ma, Xia, Xue, Wang, and Wei]{retnet}
Yutao Sun, Li~Dong, Shaohan Huang, Shuming Ma, Yuqing Xia, Jilong Xue, Jianyong Wang, and Furu Wei.
\newblock Retentive network: A successor to transformer for large language models.
\newblock \emph{arXiv preprint arXiv:2307.08621}, 2023.

\bibitem[Sun et~al.(2024)Sun, Dong, Zhu, Huang, Wang, Ma, Zhang, Wang, and Wei]{yoco}
Yutao Sun, Li~Dong, Yi~Zhu, Shaohan Huang, Wenhui Wang, Shuming Ma, Quanlu Zhang, Jianyong Wang, and Furu Wei.
\newblock You only cache once: Decoder-decoder architectures for language models.
\newblock \emph{Advances in Neural Information Processing Systems}, 37:\penalty0 7339--7361, 2024.

\bibitem[Sun et~al.(2025)Sun, Ye, Li, Xia, Chen, Gao, Cao, Wang, and Wei]{resa}
Yutao Sun, Tianzhu Ye, Dong Li, Yuqing Xia, Jian Chen, Yizhao Gao, Shijie Cao, Jianyong Wang, and Furu Wei.
\newblock Rectified sparse attention.
\newblock \emph{arXiv preprint arXiv:2506.04108}, 2025.

\bibitem[Tang et~al.(2024)Tang, Zhao, Zhu, Xiao, Kasikci, and Han]{quest}
Jiaming Tang, Yilong Zhao, Kan Zhu, Guangxuan Xiao, Baris Kasikci, and Song Han.
\newblock Quest: Query-aware sparsity for efficient long-context llm inference.
\newblock \emph{arXiv preprint arXiv:2406.10774}, 2024.

\bibitem[Team(2025)]{minicpm4}
MiniCPM Team.
\newblock Minicpm4: Ultra-efficient llms on end devices.
\newblock 2025.

\bibitem[Tillet et~al.(2019)Tillet, Kung, and Cox]{triton}
Philippe Tillet, Hsiang-Tsung Kung, and David Cox.
\newblock Triton: an intermediate language and compiler for tiled neural network computations.
\newblock In \emph{Proceedings of the 3rd ACM SIGPLAN International Workshop on Machine Learning and Programming Languages}, pages 10--19, 2019.

\bibitem[Vaswani et~al.(2017)Vaswani, Shazeer, Parmar, Uszkoreit, Jones, Gomez, Kaiser, and Polosukhin]{attention}
Ashish Vaswani, Noam Shazeer, Niki Parmar, Jakob Uszkoreit, Llion Jones, Aidan~N Gomez, {\L}ukasz Kaiser, and Illia Polosukhin.
\newblock Attention is all you need.
\newblock \emph{Advances in neural information processing systems}, 30, 2017.

\bibitem[Wang et~al.(2024)Wang, Ma, Cao, Zhang, Xue, Shi, Zheng, Miao, Yang, Cao, Yang, and Yang]{ladder}
Lei Wang, Lingxiao Ma, Shijie Cao, Quanlu Zhang, Jilong Xue, Yining Shi, Ningxin Zheng, Ziming Miao, Fan Yang, Ting Cao, Yuqing Yang, and Mao Yang.
\newblock Ladder: Enabling efficient low-precision deep learning computing through hardware-aware tensor transformation.
\newblock In \emph{18th USENIX Symposium on Operating Systems Design and Implementation (OSDI 24)}, pages 307--323, Santa Clara, CA, July 2024. USENIX Association.
\newblock ISBN 978-1-939133-40-3.
\newblock URL \url{https://www.usenix.org/conference/osdi24/presentation/wang-lei}.

\bibitem[Xiao et~al.(2024{\natexlab{a}})Xiao, Zhang, Han, Xiao, Lin, Zhang, Liu, and Sun]{xiao2024infllm}
Chaojun Xiao, Pengle Zhang, Xu~Han, Guangxuan Xiao, Yankai Lin, Zhengyan Zhang, Zhiyuan Liu, and Maosong Sun.
\newblock Infllm: Training-free long-context extrapolation for llms with an efficient context memory.
\newblock \emph{arXiv preprint arXiv:2402.04617}, 2024{\natexlab{a}}.

\bibitem[Xiao et~al.(2023)Xiao, Tian, Chen, Han, and Lewis]{streamingllm}
Guangxuan Xiao, Yuandong Tian, Beidi Chen, Song Han, and Mike Lewis.
\newblock Efficient streaming language models with attention sinks.
\newblock \emph{arXiv preprint arXiv:2309.17453}, 2023.

\bibitem[Xiao et~al.(2024{\natexlab{b}})Xiao, Tang, Zuo, Guo, Yang, Tang, Fu, and Han]{duo}
Guangxuan Xiao, Jiaming Tang, Jingwei Zuo, Junxian Guo, Shang Yang, Haotian Tang, Yao Fu, and Song Han.
\newblock Duoattention: Efficient long-context llm inference with retrieval and streaming heads.
\newblock \emph{arXiv preprint arXiv:2410.10819}, 2024{\natexlab{b}}.

\bibitem[Xu et~al.(2025)Xu, Xiao, Huang, Guo, and Han]{xu2025xattention}
Ruyi Xu, Guangxuan Xiao, Haofeng Huang, Junxian Guo, and Song Han.
\newblock Xattention: Block sparse attention with antidiagonal scoring.
\newblock \emph{arXiv preprint arXiv:2503.16428}, 2025.

\bibitem[Yang et~al.(2025{\natexlab{a}})Yang, Li, Yang, Zhang, Hui, Zheng, Yu, Gao, Huang, Lv, et~al.]{yang2025qwen3}
An~Yang, Anfeng Li, Baosong Yang, Beichen Zhang, Binyuan Hui, Bo~Zheng, Bowen Yu, Chang Gao, Chengen Huang, Chenxu Lv, et~al.
\newblock Qwen3 technical report.
\newblock \emph{arXiv preprint arXiv:2505.09388}, 2025{\natexlab{a}}.

\bibitem[Yang et~al.(2025{\natexlab{b}})Yang, Guo, Tang, Hu, Xiao, Tang, Lin, Liu, Lu, and Han]{lserve}
Shang Yang, Junxian Guo, Haotian Tang, Qinghao Hu, Guangxuan Xiao, Jiaming Tang, Yujun Lin, Zhijian Liu, Yao Lu, and Song Han.
\newblock Lserve: Efficient long-sequence llm serving with unified sparse attention.
\newblock \emph{arXiv preprint arXiv:2502.14866}, 2025{\natexlab{b}}.

\bibitem[Yang et~al.(2024{\natexlab{a}})Yang, Sheng, Gonzalez, Stoica, and Zheng]{doublesparse}
Shuo Yang, Ying Sheng, Joseph~E Gonzalez, Ion Stoica, and Lianmin Zheng.
\newblock Post-training sparse attention with double sparsity.
\newblock \emph{arXiv preprint arXiv:2408.07092}, 2024{\natexlab{a}}.

\bibitem[Yang et~al.(2023)Yang, Wang, Shen, Panda, and Kim]{gla}
Songlin Yang, Bailin Wang, Yikang Shen, Rameswar Panda, and Yoon Kim.
\newblock Gated linear attention transformers with hardware-efficient training.
\newblock \emph{arXiv preprint arXiv:2312.06635}, 2023.

\bibitem[Yang et~al.(2024{\natexlab{b}})Yang, Wang, Zhang, Shen, and Kim]{deltanet}
Songlin Yang, Bailin Wang, Yu~Zhang, Yikang Shen, and Yoon Kim.
\newblock Parallelizing linear transformers with the delta rule over sequence length.
\newblock \emph{arXiv preprint arXiv:2406.06484}, 2024{\natexlab{b}}.

\bibitem[Yuan et~al.(2025)Yuan, Gao, Dai, Luo, Zhao, Zhang, Xie, Wei, Wang, Xiao, et~al.]{nsa}
Jingyang Yuan, Huazuo Gao, Damai Dai, Junyu Luo, Liang Zhao, Zhengyan Zhang, Zhenda Xie, YX~Wei, Lean Wang, Zhiping Xiao, et~al.
\newblock Native sparse attention: Hardware-aligned and natively trainable sparse attention.
\newblock \emph{arXiv preprint arXiv:2502.11089}, 2025.

\bibitem[Zadouri et~al.(2025)Zadouri, Strauss, and Dao]{tridao_efficient_attention}
Ted Zadouri, Hubert Strauss, and Tri Dao.
\newblock Hardware-efficient attention for fast decoding.
\newblock \emph{arXiv preprint arXiv:2505.21487}, 2025.

\bibitem[Zaheer et~al.(2020)Zaheer, Guruganesh, Dubey, Ainslie, Alberti, Ontanon, Pham, Ravula, Wang, Yang, et~al.]{zaheer2020big}
Manzil Zaheer, Guru Guruganesh, Kumar~Avinava Dubey, Joshua Ainslie, Chris Alberti, Santiago Ontanon, Philip Pham, Anirudh Ravula, Qifan Wang, Li~Yang, et~al.
\newblock Big bird: Transformers for longer sequences.
\newblock \emph{Advances in neural information processing systems}, 33:\penalty0 17283--17297, 2020.

\bibitem[Zeng et~al.(2024)Zeng, Lin, Hou, Zhang, and Deng]{attentiongatekv}
Zihao Zeng, Bokai Lin, Tianqi Hou, Hao Zhang, and Zhijie Deng.
\newblock In-context kv-cache eviction for llms via attention-gate.
\newblock \emph{arXiv preprint arXiv:2410.12876}, 2024.

\bibitem[Zhang et~al.(2024)Zhang, Ji, Chen, Fu, Miao, Nie, Chen, and Cui]{pqcache}
Hailin Zhang, Xiaodong Ji, Yilin Chen, Fangcheng Fu, Xupeng Miao, Xiaonan Nie, Weipeng Chen, and Bin Cui.
\newblock Pqcache: Product quantization-based kvcache for long context llm inference.
\newblock \emph{arXiv preprint arXiv:2407.12820}, 2024.

\bibitem[Zhang et~al.(2025)Zhang, Xiang, Huang, Wei, Xi, Zhu, and Chen]{zhang2025spargeattn}
Jintao Zhang, Chendong Xiang, Haofeng Huang, Jia Wei, Haocheng Xi, Jun Zhu, and Jianfei Chen.
\newblock Spargeattn: Accurate sparse attention accelerating any model inference.
\newblock In \emph{International Conference on Machine Learning (ICML)}, 2025.

\bibitem[Zhang et~al.(2023)Zhang, Sheng, Zhou, Chen, Zheng, Cai, Song, Tian, Re, Barrett, Wang, and Chen]{h2o}
Zhenyu Zhang, Ying Sheng, Tianyi Zhou, Tianlong Chen, Lianmin Zheng, Ruisi Cai, Zhao Song, Yuandong Tian, Christopher Re, Clark Barrett, Zhangyang Wang, and Beidi Chen.
\newblock H2o: Heavy-hitter oracle for efficient generative inference of large language models.
\newblock In \emph{Thirty-seventh Conference on Neural Information Processing Systems}, 2023.
\newblock URL \url{https://openreview.net/forum?id=RkRrPp7GKO}.

\bibitem[Zheng et~al.(2024)Zheng, Yin, Xie, Sun, Huang, Yu, Cao, Kozyrakis, Stoica, Gonzalez, et~al.]{sglang}
Lianmin Zheng, Liangsheng Yin, Zhiqiang Xie, Chuyue~Livia Sun, Jeff Huang, Cody~Hao Yu, Shiyi Cao, Christos Kozyrakis, Ion Stoica, Joseph~E Gonzalez, et~al.
\newblock Sglang: Efficient execution of structured language model programs.
\newblock \emph{Advances in Neural Information Processing Systems}, 37:\penalty0 62557--62583, 2024.

\bibitem[Zhu et~al.(2022)Zhu, Wu, Diao, Ke, Li, Zhang, Xue, Ma, Xia, Cui, Yang, Yang, Zhou, Cidon, and Pekhimenko]{roller}
Hongyu Zhu, Ruofan Wu, Yijia Diao, Shanbin Ke, Haoyu Li, Chen Zhang, Jilong Xue, Lingxiao Ma, Yuqing Xia, Wei Cui, Fan Yang, Mao Yang, Lidong Zhou, Asaf Cidon, and Gennady Pekhimenko.
\newblock {ROLLER}: Fast and efficient tensor compilation for deep learning.
\newblock In \emph{16th USENIX Symposium on Operating Systems Design and Implementation (OSDI 22)}, pages 233--248, Carlsbad, CA, July 2022. USENIX Association.
\newblock ISBN 978-1-939133-28-1.
\newblock URL \url{https://www.usenix.org/conference/osdi22/presentation/zhu}.

\end{thebibliography}
\newpage
\appendix


\end{document}